\documentclass[lettersize,journal]{IEEEtran}

\usepackage{graphicx}
\usepackage{booktabs}
\usepackage[linesnumbered,vlined,ruled]{algorithm2e}
\usepackage{setspace}
\newcommand{\funcfont}[1]{{\fontfamily{qcr}\selectfont\small $\thinspace$#1$\thinspace$}}
\newcommand{\appropto}{\mathrel{\vcenter{
  \offinterlineskip\halign{\hfil$##$\cr
    \propto\cr\noalign{\kern2pt}\sim\cr\noalign{\kern-2pt}}}}}
\usepackage[accsupp]{axessibility}  
\usepackage{tabularray}
\usepackage{tabularx}

\usepackage[pagebackref,breaklinks,colorlinks]{hyperref}

\usepackage{orcidlink}
\usepackage{multicol,multirow,booktabs,mathabx}

\let\oldnl\nl
\newcommand{\nonl}{\renewcommand{\nl}{\let\nl\oldnl}}

\usepackage{colortbl}
\definecolor{mygray}{gray}{.95}

\usepackage{pifont}
\newcommand{\cmark}{\ding{51}}%
\newcommand{\xmark}{\ding{55}}%

\makeatletter
\DeclareRobustCommand\onedot{\futurelet\@let@token\@onedot}
\def\@onedot{\ifx\@let@token.\else.\null\fi\xspace}

\makeatother

\begin{document}

\title{Efficient 3D Multi-Camera Multi-Object Tracking and Pose Estimation}

\author{Linh~Van~Ma, Tran~Thien~Dat~Nguyen, Juhua~Hu, Wei~Cheng, Moongu~Jeon,~\IEEEmembership{Senior Member,~IEEE}
\IEEEcompsocitemizethanks{\IEEEcompsocthanksitem Linh Van Ma and Moongu Jeon are with the School of Electrical Engineering and Computer Science at GIST, Gwangju, Korea (e-mail: \{linh.mavan, mgjeon\}@gist.ac.kr). Tran Thien Dat Nguyen is with the School of Electrical Engineering, Computing and Mathematical Sciences, Curtin University, Australia (e-mail: t.nguyen1@curtin.edu.au). Juhua Hu, Wei Cheng are with the School of Engineering and Technology, University of Washington, USA (e-mail: \{juhuah,uwcheng\}@uw.edu).\\The authors would like to acknowledge Professor Ba-Ngu Vo (ba-ngu.vo@curtin.edu.au ) for his invaluable discussions during the drafting of this manuscript. }}


\maketitle

\begin{abstract}


This paper proposes a fast and online method for jointly performing 3D multi-object tracking and pose estimation using multiple monocular cameras. Our algorithm requires only 2D bounding box and pose detections, eliminating the need for costly 3D training data or computationally expensive deep learning models. Our solution is an efficient implementation of a Bayes-optimal multi-object tracking filter, enhancing computational efficiency while maintaining accuracy. We demonstrate that our algorithm is significantly faster than state-of-the-art methods without compromising accuracy, using only publicly available pre-trained 2D detection models. We also illustrate the robust performance of our algorithm in scenarios where multiple cameras are intermittently disconnected or reconnected during operation.
\end{abstract}


   
\section{Introduction}
\label{sec:intro}

Tracking multiple objects and their poses in 3D using only cameras is an important problem in computer vision. This challenge hosts a range of applications, including sports analytics, age care, surveillance, and human-machine interaction \cite{bridgeman2019multi,bradler2021urban}. Recently, learning-based algorithms have demonstrated their effectiveness in estimating object 3D locations and poses. However, as outlined in \cite{liao2024multiple}, the prior knowledge acquired by these 3D models may not generalize well to scenarios outside their training datasets. Further, the process of labeling 3D training data is labor-intensive, and large models are often necessary for effective learning, which can extend inference times and render these 3D learning-based methods less suitable for real-time applications.

In contrast, 2D object detection and pose estimation have been effectively addressed in the literature. Compared to 3D models, 2D models require less memory and offer faster training and inference times. Additionally, the availability of 2D training data is currently greater and easier to obtain than 3D data, leading to more robust and generalizable algorithms. Well-known 2D detection algorithms such as YOLO \cite{ge2021yolox}, OpenPose \cite{cao2017realtime}, and AlphaPose \cite{fang2022alphapose} can detect humans and their poses in real-time with a high level of accuracy. 

Leveraging the advancement of 2D object and pose detection, a promising approach for real-time 3D multi-object tracking and pose estimation involves fusing 2D detections from multiple cameras to create 3D trajectories of poses. The algorithms presented in \cite{ong2020bayesian,linh2024inffus} use 2D detection from various cameras to estimate 3D trajectories. These techniques are based on the multi-sensor generalized labeled multi-Bernoulli (MS-GLMB) filter \cite{vo2019multi}, a Bayes-optimal filter designed for multi-object estimation. However, due to their consideration of multiple data association hypotheses, these methods can be computationally intensive, which limits their capability for real-time tracking. Further, they only estimate 3D trajectories and shapes, excluding 3D object poses in their estimation schemes.

Our work is motivated by the need for real-time 3D multi-object tracking and pose estimation. Our approach relies solely on 2D object and pose detection. We consider the correspondence between 2D views as a multi-sensor multi-object estimation problem and  
address this challenge using the Bayes-optimal MS-GLMB filter. To reduce the computational costs, we propose a series of approximations to the filter, making real-time processing feasible. In summary, our contributions are:
\begin{itemize}
    \item A fast 3D algorithm for jointly estimating multi-object poses and trajectories that requires only 2D object bounding box and pose detections, designed for real-time applications.
    \item The derivation of the proposed algorithm from Bayes-optimal multi-object tracking filter perspective.
    \item A comprehensive series of studies demonstrating the algorithm’s performance and robustness across various datasets and configurations.
\end{itemize}

The rest of the paper is organized as follows. Section \ref{sec:backgrounds_sfmcmt} presents a brief overview of related works in 2D and 3D visual object detection, tracking, and pose estimation. In Section \ref{sec:Algorithm}, we propose an efficient algorithm for 3D multi-camera multi-object tracking and pose estimation using 2D detection. In Section \ref{sec:alg-details}, we detail the implementation of the proposed algorithm. Section \ref{subsec:Relevance-MSGLMB} provides the rationale for our algorithm, which is an approximation of the Bayes-optimal MS-GLMB filter. Finally, the experimental results on various tracking scenarios and two ablation studies are presented in Section \ref{sec:experiment_sfmcmt} while Section \ref{conclusion} concludes the paper.

\section{Related Works}\label{sec:backgrounds_sfmcmt}

\subsection{3D Multi-Camera Multi-Object Tracking}
Early approaches in 3D object tracking use multiple cameras to track the head locations of pedestrians using homography transformation \cite{khan2006multiview,eshel2008homography}. The approaches in \cite{chavdarova2017deep,baque2017deep} use deep-learning-based 3D detectors and a tracker to estimate 3D tracks. The recent approach in \cite{nguyen2022lmgp} utilizes a spatial-temporal lifted multi-cut formulation to track 3D objects, whereas \cite{teepe2024earlybird} employs ground plane (birds-eye) information for early fusion to improve 3D tracking accuracy. However, these approaches only estimate object ground plane locations, ignoring object 3D shapes and poses. The method in \cite{gao2023delving}, while producing 3D trajectories, requires 3D detections. Nevertheless, training and inferring the 3D object detection/tracking model is computationally demanding, and most of the trained 3D models need to be retrained when the camera geometry changes to achieve good performance.  Hence, its applicability is limited in real-time applications. Recently, the online multi-camera multi-object tracking algorithm in \cite{shim2023fast} efficiently clusters 2D tracks from different cameras using hierarchical clustering, but the resultant tracks are not in 3D. An alternative is directly performing 3D tracking using 2D detection, in which the camera geometry is integrated into the tracker. The filters proposed in \cite{ong2020bayesian,linh2024inffus}, based on the MS-GLMB filter \cite{vo2019multi}, follow such a pathway. However, their computation time is relatively high and not suitable for real-time tracking with a large number of objects.

\subsection{3D Multi-Object Pose Estimation} 
Earlier works in 3D pose estimation modeled poses as graphical structures, where nodes represent joints and edges depict their relationships \cite{belagiannis20143d}. 3D poses are estimated using maximum a posteriori (MAP) methods, incorporating 2D joint detections and physical constraints. More recent approaches follow a two-step process: first, localizing humans in 3D and then estimating detailed 3D poses. For example, VoxelPose \cite{tu2020voxelpose} uses two networks: one for localizing 3D proposals, and another for estimating the detailed 3D poses, claiming robustness to varying camera views. Meanwhile, Faster-VoxelPose \cite{ye2022faster} eliminates the 3D networks, resulting in a tenfold increase in speed. TesseTrack \cite{reddy2021tessetrack} offers an end-to-end learnable framework that detects persons in 3D and computes temporal features from neighboring frames for pose estimation. The method in \cite{wu2021graph} employs graph neural networks for 3D localization, followed by coarse and fine 3D pose estimation using relational information among joints and views. Conversely, MvP \cite{zhang2021direct} directly learns poses from multi-view images using multi-hierarchical query embeddings and projective attention to aggregate complementary information. SelfPose3D \cite{srivastav2024selfpose3d} generates 3D joints from 2D affine augmentations without requiring 3D training data; it uses augmented images with VoxelPose to compute loss against 2D pseudo poses obtained from a detection network. On the other hand, the DAS algorithm in \cite{wang2022distribution} is designed to learn the 2D to 3D mapping, and represents the 3D joint distribution by normalization flow models, where the joint locations are obtained via a residual log-likelihood estimator.

Conversely, some methods focus on geometry-based techniques for 2D view correspondence. For instance, \cite{lin2021multi} achieves cross-view correspondence and 3D reconstruction using a plane sweep method, projecting 2D poses to a virtual depth plane and refining them based on cross-view consistency. This involves two levels of depth regression: one at the person level and another at the joint level, both trained on synthetic 3D data. The method in \cite{liao2024multiple} generalizes to new camera geometries by integrating a learning-free module for handling 3D geometry with a learnable appearance module for end-to-end 2D pose estimation. MVPose \cite{dong2019fast} employs a 2D detector for bounding box detection and utilizes appearance feature cues and geometric pose consistency to compute association scores through iterative optimization. \cite{chen2020multi} critiques methods based on epipolar constraints, arguing that they can misalign human joints; instead, they match human feet across frames to construct joint positions through MAP estimation, addressing detection inaccuracies in occluded scenes. 



\subsection{2D Multi-Object Detection and Pose Estimation}
The  2D object detection and pose estimation problems have been studied for decades. Early 2D detectors based on hand-crafted features \cite{lowe2004distinctive,dalal2005histograms} were popular. Later, the idea of using convolutional neural networks (CNN) \cite{girshick2014rich} to extract features from images for object detection has dominated the literature due to CNN's capability to capture multi-scale representations. Popular CNN-based detectors \cite{girshick2015fast,ren2015faster} use region proposals to obtain detections. However, its efficiency is surpassed by the anchor-free detectors \cite{redmon2016you}. The recent anchor-free YOLOX algorithm \cite{ge2021yolox} can detect objects in real-time with a high level of accuracy. 

Detecting human poses from 2D images is typically addressed through two main approaches. The bottom-up approach detects all body parts and subsequently associates them with individuals, where human parts are identified using learning-based methods, and poses are represented as graphical models, excelling in handling occlusions \cite{chen2015parsing}. The rapid OpenPose method \cite{cao2017realtime} utilizes part affinity fields to assess the degree of association between parts and individuals, framing the association as a matching problem, while OpenPifpaf \cite{kreiss2021openpifpaf} employs composite fields (part intensity and part association) to locate and connect human parts effectively. Conversely, the top-down approach performs single-person pose estimation for each detected bounding box around the human body, with models like Region-based CNN, Feature Pyramid Networks, and Residual Networks adapted for pose estimation \cite{he2017mask}, \cite{chen2018cascaded}, \cite{xiao2018simple}, and a multi-scale feature network proposed for pose estimation in \cite{sun2019deep}. Recently, AlphaPose \cite{fang2022alphapose} claimed to be the first framework that performs unified 2D whole-body pose estimation and tracking.

\section{The Fast 3D Multi-Object Tracking and Pose Estimation Algorithm\label{sec:Algorithm}}

\begin{figure}[tbh]
\begin{centering}
\includegraphics[width=0.8\columnwidth]{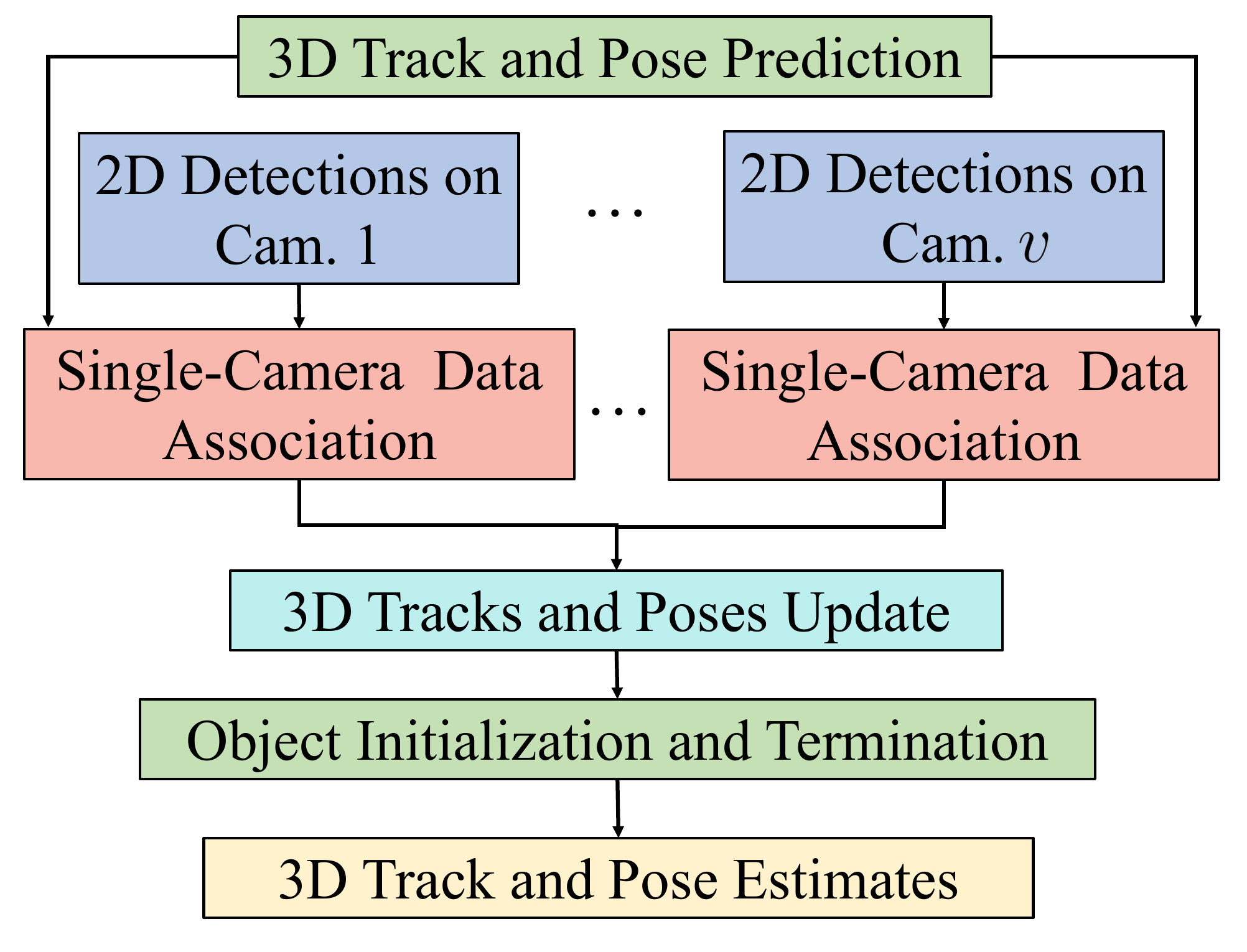}
\par\end{centering}
\caption{Overview of the proposed algorithm with $\upsilon$ cameras.\label{fig:Overview-algo}}
\end{figure}

This section presents the proposed 3D multi-object tracking and pose estimation algorithm. Our method is based on Bayes-optimal data fusion to estimate 3D object states from 2D object and pose detections obtained from multiple cameras. In particular, we propose a drastic approximation to the multi-sensor generalized labeled multi-Bernoulli (MS-GLMB) filter, which substantially improves processing speed. Intuitively, the resulting filtering recursion can be broken into modules, including: track prediction; data association and track update; track initialization and termination. An overview of the algorithm implementation is illustrated in Fig. \ref{fig:Overview-algo}. Since our 3D tracking process does not involve training or inferring a deep learning model, the tracker requires no graphical processing unit (GPU) acceleration and achieves efficient computation by only processing the tracks and solving a series of 2D linear assignment problems. Nonetheless, the algorithms that detect 2D bounding boxes and poses may require training on 2D data, although our technique is not restricted to any specific type of 2D detectors.

\subsection{Track Prediction}

In our algorithm, each object (track) is represented by a distinct ID $\ell$ that is unchanged over time and a state $x$ contains (real-world) 3D position, velocity,
shape parameters, and keypoints (joint locations) and their velocity in 3D. Using the notation $1:n$ for the list $1,2,...,n$,
the state vector $x$ can be expressed as 
\[
x=[\rho^{(1:3)},\dot{\rho}^{(1:3)},s^{(1:3)},p^{(1:3P)},\dot{p}^{(1:3P)}]^{T},
\]
where $\rho^{(1:3)}$ is the 3D position of the object center, $\dot{\rho}^{(1:3)}$ is
the 3D velocity of the object, $s^{(1:3)}$ is the shape parameters of the 3D axis-aligned ellipsoid enclosing the object, which is the logarithm of the half-lengths of the principal axes, and $p^{(1:3P)}$ is the object 3D keypoints with $P$ is the (fixed) total number of keypoints and $\dot{p}^{(1:3P)}$ are their corresponding 3D velocities.  Our proposed method applies to all keypoint conventions, such as the \textit{BODY\_25}   ($P=25$) keypoint convention introduced in \cite{cao2017realtime}, COCO (18 keypoints) \cite{lin2014microsoft}, and MPII (15 keypoints) \cite{andriluka20142d}. When the object first appears, we assume its state is Gaussian distributed, represented by a mean $\mu_{0}^{(\ell)}$ and a covariance matrix $\mathbf{P}_{0}^{(\ell)}$, i.e., $\mathcal{N}\left(x;\mu_{0}^{(\ell)},\mathbf{P}_{0}^{(\ell)}\right)$.

To predict the object state from time step $t-1$ to $t$, we assume the object motion follows the constant velocity model, a popular motion model used in visual tracking when the frame rate is relatively high. Since an object's size remains mostly unchanged over time, we set the mean of the object shape parameters at the next time step to be the same as at the current time (i.e., a random walk model on the object shape state). The 3D keypoints also follow the constant velocity model. However, since the keypoints could change drastically when the object changes their pose, we set the noise variance of the keypoints to be relatively high. Overall, the transition of the object state from one time step to the next is linear. If the probability density of the object state with ID $\ell$ at $t-1$ is a Gaussian with mean $\mu_{t-1}^{(\ell)}$ and covariance matrix $\mathbf{P}_{t-1}^{(\ell)}$, the predicted mean and covariance matrix of the (Gaussian) object state density at $t$ is computed by the well-known Kalman prediction \cite{ong2020bayesian}, yielding the predicted mean and covariance matrix $\mu_{t|t-1}^{(\ell)}$ and $\mathbf{P}_{t|t-1}^{(\ell)}$, respectively.

\subsection{Data Association and Track Update}
\begin{figure}[h!]
\begin{centering}
\includegraphics[width=0.52\columnwidth]{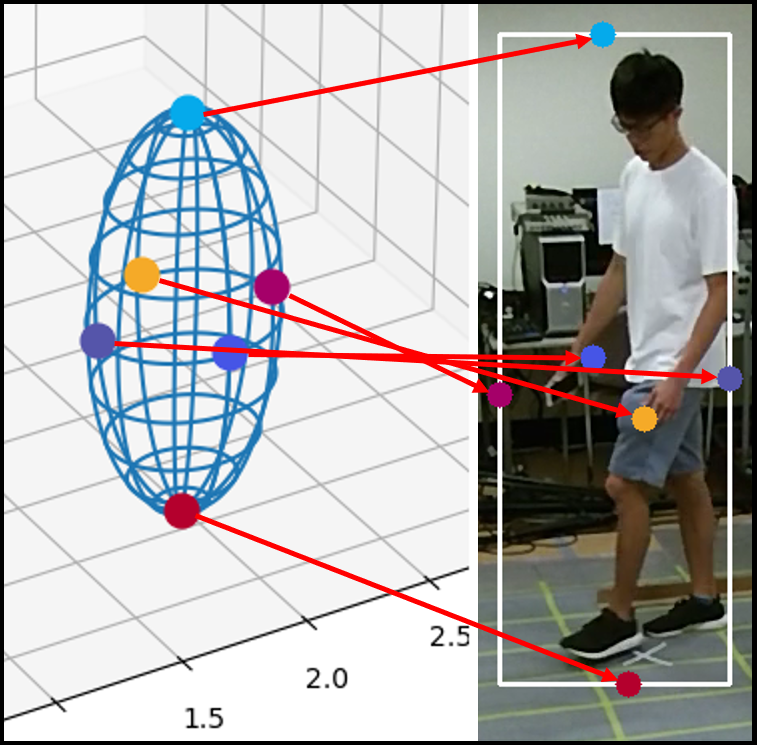}
\par\end{centering}
\caption{$\Phi^{(c)}_{b}(\cdot)$ first projects selected extreme points on 3D ellipsoid to 2D camera plane. The projected 2D bounding box is formed by taking the maximum/minimum of the coordinates (reference to the 2D camera plane) of the projected 2D points. \label{fig:ukf_example}} 
\end{figure}
Suppose that a set of monocular cameras, indexed from $1$ to $v$,
are used to detect objects and their poses in the scene. An object with state $x$ and ID $\ell$, a labeled state $(x,\ell)$,
can either be miss-detected or generate a bounding box detection $b$ associated with a keypoint list $k=[k^{(1)},...,k^{(P)}]$. If the $i^{th}$ keypoint is miss-detected, conventionally, $k^{(i)}=\overline{\infty}$ (a vector whose components are $\infty$). The likelihood of a track $(x,\ell)$ generating a detection $(b,k)$ on camera $c$ is given by
\begin{multline}
g^{(c)}(b,k|x,\ell)=\mathcal{N}\left(b;\Phi^{(c)}_{b}(x),\mathbf{R}_{b}^{(c)}\right) \times \\  \prod _{i=1}^{P}\left[\left(1-\delta_{\overline{\infty}}[k^{(i)}]\right)\mathcal{N}\left(k^{(i)};\Phi^{(c)}_{k,i}(x),\mathbf{R}^{(c)}_{k}\right)+\delta_{\overline{\infty}}[k^{(i)}] \right],\label{eq:sc-likelihood}
\end{multline}
where $\Phi^{(c)}_{b}(\cdot)$ and $\Phi^{(c)}_{k,i}(\cdot)$ map the 3D object state to, respectively, the corresponding
2D bounding box and the $i^{th}$ element of the 2D keypoint detection list,
$\mathbf{R}^{(c)}_{b}=\textrm{diag}\left(\left[\sigma_{\rho}^{(c)},\sigma_{s}^{(c)}\right]\right)$ with
$\sigma_{\rho}^{(c)}$ and $\sigma_{s}^{(c)}$ are the variances
of the position and size of the 2D bounding box detection noise, respectively, $\mathbf{R}^{(c)}_{k}=\textrm{diag}\left(\sigma_{k}^{(c)}\right)$ with $\sigma_{k}^{(c)}$ is the variance of the 2D keypoints detection noise, and $\delta_{x}[y]$ is the Kronecker delta function that takes 1 if $x=y$ and 0 otherwise. The projection $\Phi^{(c)}_{b}(\cdot)$ from 3D position (represented by a 3D ellipsoid) to 2D points is obtained by mapping the extreme points of the 3D ellipsoid to their corresponding points on camera plane \cite{zhang2000flexible}. This projection is illustrated in Fig. \ref{fig:ukf_example}. The projection from the $i^{th}$ 3D state to the $i^{th}$ 2D keypoint element, $\Phi^{(c)}_{k,i}(\cdot)$, is performed using the homogeneous projection method. 

\emph{Remark:} Our models can also handle object appearance features, similar to ones in \cite{linh2024inffus}. But for efficient computation, we omit these features in this work. Nevertheless, we note that although appearance features can help re-identify objects during missed-detection events, unreliable features may distort the assignment scores and lead to incorrect data associations.

Since object keypoints vary significantly between frames, the model of the 3D keypoint positions is uninformative, as mentioned in the previous subsection. Thus, we omit the keypoints when computing the cost function for data association. At time $t$, we consider a set of $n$ labeled objects $\{(x^{(1)},\ell^{(1)}),...,(x^{(n)},\ell^{(n)})\}$.
For each camera, say camera $c$, with a set of enumerated 2D bounding box and keypoint detections
$B^{(c)}=\{(b_{1}^{(c)},k_{1}^{(c)}),...,(b_{m}^{(c)},k_{m}^{(c)})\}$ (which may contain detections generated by objects and false positive detections), we define a mapping
$\gamma^{(c)}$ that maps the object IDs to the index of the detections
in $B^{(c)}$. $\gamma^{(c)}(\ell)=i>0$ means the $i^{th}$ detection in $B^{(c)}$ is assigned to the object, and $\gamma^{(c)}(\ell)=0$
means no detection from camera $c$ is assigned to the object. We
obtain the optimal association map $\gamma^{(c)}$ by solving the
following optimization problem
\begin{equation}
\gamma^{*(c)}=\underset{\gamma^{(c)}}{\arg\min}\sum_{\ell\in\{\ell^{(1)},..,\ell^{(n)}\}}C^{(c)}(\ell,\gamma^{(c)}(\ell)),\label{eq:cost-per-cam}
\end{equation}where: 
\begin{align}
\!\!\!C^{(c)}\!(\ell,j) & \!=\!-\left(1-\delta_{j}[0]\right)\log\left(q^{(\ell,c)}(b_{j}^{(c)})\right)+\iota_{\infty}\delta_{j}[0];\label{eq:match-cost}\\
q^{(\ell,c)}(b) & \!=\!\int \!\mathcal{N} \left(x;\mu_{t|t-1}^{(\ell)}, \mathbf{P}_{t|t-1}^{(\ell)}\right) \mathcal{N}\left(b;\Phi^{(c)}_{b}(x),\mathbf{R}^{(c)}_{b}\right)\!dx;\label{eq:q-per-cam}
\end{align}
$\iota_{\infty}$ is a large number ($\exp(-\iota_{\infty})\approx0$) (i.e., assume
the miss-detection cost is much higher than other assignment costs). The integral in (\ref{eq:q-per-cam}) can be computed approximately with unscented transform \cite{julier2004unscented,van2004sigma} such that
\begin{equation}
q^{(\ell,c)}(b)\approx\mathcal{N}\left(b;\mu_{u}^{(\ell,c)},\mathbf{R}^{(c)}_{b}+\mathbf{P}_{u}^{(\ell,c)}\right),\label{eq:q_z}
\end{equation} where \begin{align}
\mu_{u}^{(\ell,c)} & =\sum_{i=0}^{2L}w_{m,i}^{(\ell)}\mathcal{Y}_{i}^{(\ell,c)},\\
\mathbf{P}_{u}^{(\ell,c)} & =\sum_{i=0}^{2L}w_{c,i}^{(\ell)}(\mathcal{Y}_{i}^{(\ell,c)}\!-\!\mu_{u}^{(\ell,c)})(\mathcal{Y}_{i}^{(\ell,c)}\!-\!\mu_{u}^{(\ell,c)})^{T},
\end{align}\vspace{-1em}\begin{align}
\mathcal{Y}_{i}^{(\ell,c)} & =\Phi^{(c)}(\mathcal{X}_{i}^{(\ell)}),\\
(\mathcal{X}_{i}^{(\ell)},w_{m,i}^{(\ell)},w_{c,i}^{(\ell)}) & =\mathcal{U}(\mu_{t|t-1}^{(\ell)},\mathbf{P}_{t|t-1}^{(\ell)}),i=0:2L,
\end{align}
$\mathcal{U}(\cdot)$ generates $2L+1$  sigma points from the unscented
transform given the Gaussian
mean and covariance matrix, $L$ is the dimension of the state vector, $(\mathcal{X},w_{m},w_{c})$ is tuple
of a sigma point and its corresponding mean and covariance weights,
respectively, and $\mu_{t|t-1}^{(\ell)}$ and $\mathbf{P}_{t|t-1}^{(\ell)}$
are respectively the mean and covariance matrix of the predicted density of object with ID $\ell$.  The rationale of the optimization given in (\ref{eq:cost-per-cam}) follows from the approximation of the MS-GLMB filter, which is detailed in Section \ref{subsec:Relevance-MSGLMB}.

The optimization in (\ref{eq:cost-per-cam}) can be solved efficiently with a standard
2D linear assignment algorithm, e.g., Hungarian \cite{kuhn1955hungarian}
or Jonker-Volgenant \cite{jonker1987shortest}. We use detection gating to eliminate the unlikely solutions. In particular,
for object $\ell$ with mean state $\mu^{(\ell)}$ and the $j^{th}$ detection $b_j$ from
camera $c$, let $\mathcal{G}_{s}(\mu^{(\ell)})$ be the mean ground plane position
of $\ell$ and $\mathcal{G}_{d}^{(c)}(b_j)$ be the ground plane position
of the projection of $b_j$ on 3D real-world coordinate, if $\left\Vert \mathcal{G}_{s}(\mu^{(\ell)})-\mathcal{G}_{d}^{(c)}(b_j)\right\Vert _{2}>\tau_{G}$,
we set $C^{(c)}(\ell,j)=\infty$ (i.e., assigning $b_j$ to $\ell$ is infeasible). Further, we apply detection gating directly on the cost value such that if $C^{(c)}(\ell,j)>\tau_{C}$ then $C^{(c)}(\ell,j)=\infty$. The optimal multi-camera data association map is the combination of the optimal single-camera data association maps, i.e., $\gamma^{*}=[\gamma^{*(1)},...,\gamma^{*(v)}]$.
\begin{figure}
\begin{centering}
\includegraphics[width=0.95\columnwidth]{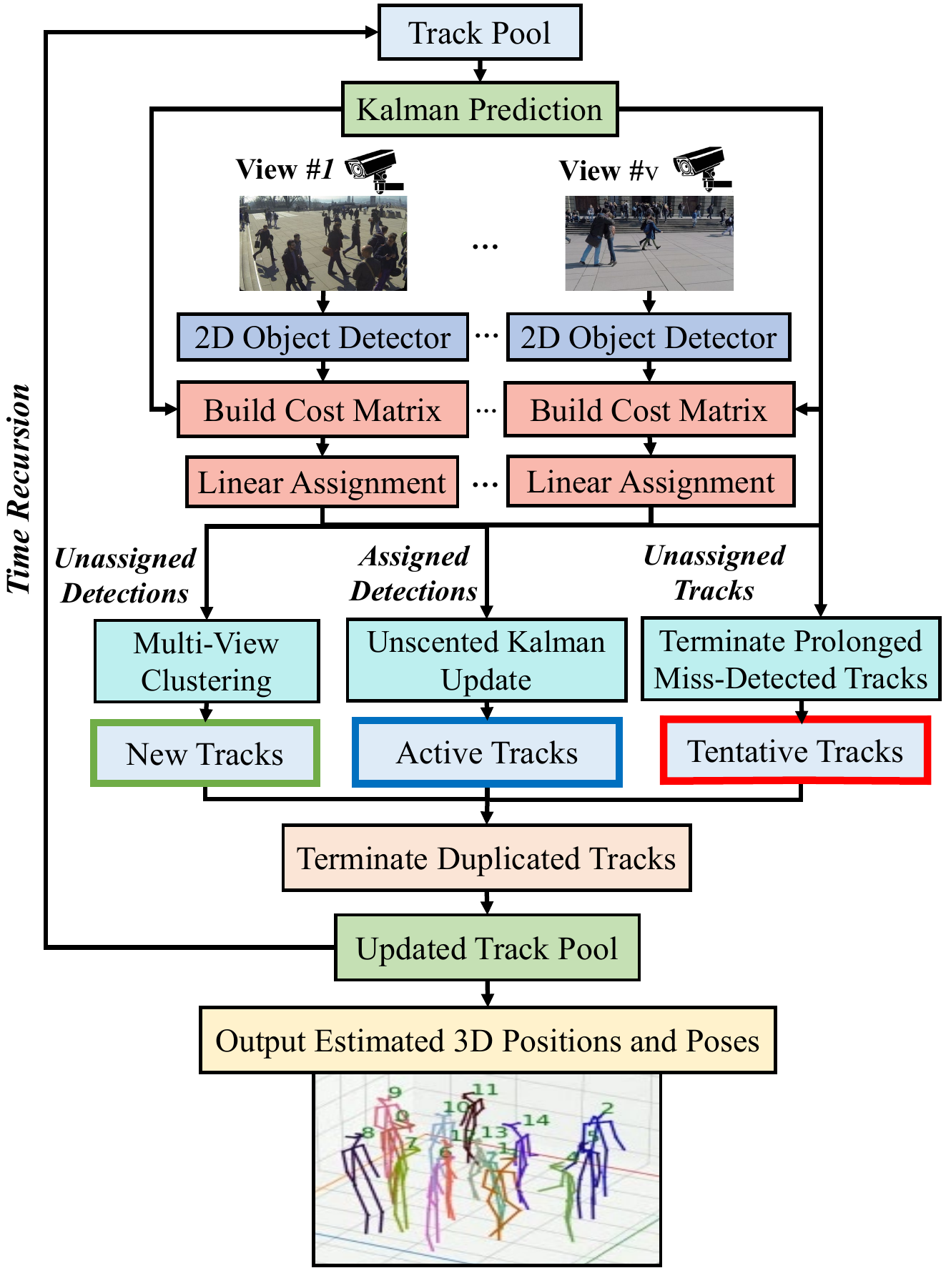}
\par\end{centering}
\caption{Algorithm flowchart. Tracks (3D kinematics, shapes, and poses) in the pool are predicted forward using Kalman prediction. For each camera, a cost matrix to assign predicted tracks to 2D bounding box detections is computed, and the optimal data association is determined. Tracks are updated once for each camera (with assigned detection) in turn using an unscented Kalman update. Unassigned detections are clustered in groups with a mean-shift clustering algorithm, and each group is used to initialize a new track. Unassigned tracks that are miss-detected for a long time are terminated. Duplicated tracks (determined via checking pair-wise similarity) are also terminated. Output 3D position and pose are the mean values of the Gaussian state probability density of new and active tracks. \label{fig:algflowchart}}
\end{figure}
Given $\gamma^{*}$, we use unscented Kalman update \cite{julier2004unscented,van2004sigma}
to update the predicted mean and covariance matrix of the object state density (3D position, velocity, shapes, keypoints, and keypoint velocity), once for each camera in turn (we skip a camera if the object is miss-detected on that camera). If no detection is assigned to the object at all, the updated
mean and covariance matrix are the same as the predicted
ones. 


\subsection{Track Initialization and Termination}

We utilize unassigned detections to initialize new tracks, recalling that the initial state probability density of a track follows a Gaussian distribution. For a camera \textit{$c$}, we apply the transformation $\bar{\Phi}^{(c)}_{b}(\cdot)$ to convert the center of the bottom edge of a 2D detected bounding box into its corresponding location on the ground plane in 3D real-world coordinates. To group the 3D projected detections from various cameras, we employ the mean shift clustering method proposed in \cite{linh2024inffus}. The centroids of these groups serve as the means ground plane location for the initial track densities. We assume standard values for the mean shape parameter of the object (representing an average human size) and set the initial mean velocity to zero. The initial covariance matrix is defined as a diagonal matrix with high variance for state components with significant initialization uncertainty. For the 3D keypoints, we first initialize a Gaussian distribution centered on the keypoint positions that correspond to a standard pose (standing upright). Subsequently, we utilize the grouped 2D pose detections identified by the mean shift clustering algorithm to refine the initial distribution, using this updated distribution for the keypoint components in the initial state probability density. 

The collection of all tracks is referred to as a \textit{track pool}\textcolor{red}{}.
After track initialization, our track pool has three track types: 
\begin{itemize}
\item New tracks - tracks that are initialized from current unassigned detection(s); 
\item Active tracks - tracks that currently have detection(s) assigned to
them; 
\item Tentative tracks - tracks that currently do not have any detection
assigned to them. 
\end{itemize}
The track termination process follows ByteTrack \cite{zhang2022bytetrack}
and FairMOT \cite{zhang2021fairmot} but replacing the 2D IoU similarity
measure (for deleting duplicated tracks) with a 3D IoU measure. After
track termination, the set of track estimates (output of the tracker)
are the mean states of the remaining (not terminated) new
and active tracks. The algorithm flowchart is given in Fig. \ref{fig:algflowchart}.

\emph{Remark:} While the proposed algorithm only selects one data association hypothesis, tracks that do not appear in this hypothesis (e.g., due to miss-detection) are not immediately discarded. Instead, they are stored in the track pool as tentative tracks, enabling revision if they are re-detected later. Thus, this feature prevents the tracker from frequently switching object IDs or incorrectly terminating tracks even when the false negative rate is relatively high.

\section{Algorithm Implementation} \label{sec:alg-details}
In this section, we provide the pseudocode for implementing our estimation technique in  Algorithm \ref{alg:sf3dmot}.  Details of the sub-routines can be found in  Algorithms \ref{alg:ukfupdatepersensor}-\ref{alg:genmsobservation}, Appendix A.  Each track is represented by a tuple $(\ell,\mu,\mathbf{P})$, where $\ell$ is the object's unique ID, $\mu$ is the mean and $\mathbf{P}$ is the covariance matrix of the state density. Detection is an array of lists $B=[B^{(1)},...,B^{(v)}]$, where  $B^{(c)}=\{(b_1,k_1),...,(b_{m^{(c)}},k_{m^{(c)}})\}$, $m^{(c)}$ is the number of 2D detections from camera $c$. Each element $b_j$ is a 2D bounding box with the format $\big[\text{left, top, log(width), log(height)}\big]^T$ and each $k_j$ is a 2D keypoint detection list. Note that if the $i^{th}$ keypoint of $k_j$ is miss-detected, conventionally, $k_j^{(i)}=\overline{\infty}$. Given the track pool at time $t-1$ with $n$ objects $\mathcal{P}_{t-1}=\{(\ell_i,\mu_i,\mathbf{P}_i)\}_{i=1:n}$ (a list of $n$ tuples), and the detection $B_t$, the track pool at $t$ and the 3D track estimates are computed using Algorithm \ref{alg:sf3dmot}. We use $\mu_i^{(s)}$ and $\mathbf{P}_i^{(s)}$ for the mean and covariance relating to the kinematic (position and velocity) and shape of the track (lines 12 and 25 of Algorithm \ref{alg:sf3dmot}), respectively. We use $\mu_i^{(p)}$ and $\mathbf{P}_i^{(p)}$ for the mean and covariance relating to the keypoints of the track (line 26 of Algorithm \ref{alg:sf3dmot}), respectively. 

\begin{algorithm}
\SetInd{0.4em}{0.5em}
\setstretch{0.9}
\fontsize{10pt}{10pt}
\DontPrintSemicolon
\nonl \textbf{Input}:  Track pool at $t$-1:$\;\mathcal{P}_{t\!-\!1}\!=\!\{(\ell_i,\mu_i,\mathbf{P}_i)\}_{i=1:n}$.\\
\nonl 2D bounding box and pose detections at $k$: $B_t=[B^{(1)},...,B^{(v)}]$, $B^{(c)}=\{(b_1,k_1),...,(b_{m^{(c)}},k_{m^{(c)}})\}$. \;
\nonl \textbf{Output}: 3DPositions{\&}Poses, $\mathcal{P}_t$.
\rule[0.5pt]{0.925\columnwidth}{0.1pt} \;
\funcfont{/*Track Prediction*/}\;
\For{$i=1:n$}{
    $(\mu_i, \mathbf{P}_{i})$$\leftarrow$\funcfont{KalmanPrediction}($\mu_i, \mathbf{P}_{i}$) 
}

\funcfont{/*Data Association*/}\;
$A\leftarrow\mathbf{0}_{n\times v}$\;
\For{$c=1:v$}{
    $C\leftarrow\mathbf{0}_{n\times m^{(c)}}$\;
    \For{$i=1:n$}{
        \For{$j=1:m^{(c)}$}{
            $s\leftarrow$\funcfont{DetectionGating}$(b_j, \mu_i^{(s)}, c)$ (Algorithms \ref{alg:det-gating} and \ref{alg:2DtoGP}, Appendix A)\;
            \If{s is true}{
                $(q, \_, \_)\leftarrow$\funcfont{UKFUpdateKS}($b_j, \mu_i^{(s)}, \mathbf{P}_{i}^{(s)}, c$) (Algorithms \ref{alg:ukfupdatepersensor}, \ref{alg:ut}, and \ref{alg:genmsobservation}, Appendix A)\;
                $C[i, j] \leftarrow -\text{log}(q)$ \;
                \If{$C[i, j]>\tau_C$}{
                    $C[i, j] \leftarrow \infty$
                }
            }
            \Else{
                $C[i, j] \leftarrow \infty$ \;
            }
        }
    }
    $A[:,c]$$\leftarrow$\funcfont{2DLinearAssignment}($C$) \;
}

\funcfont{/*Track Update*/} \;
$\mathcal{A}\leftarrow$\funcfont{EmptyList} \quad\funcfont{/* Active Tracks */}\;
\For{$i=1:n$}{
    \For{$c=1:v$}{
        \If{$A[i,c]>0$}{
            $j=A[i,c]$\;           $(\_,\mu_i^{(s)},\mathbf{P}_{i}^{(s)})\leftarrow $\funcfont{UKFUpdateKS}($b_j,\mu_i^{(s)}, \mathbf{P}_{i}^{(s)}\!,c$)  (Algorithms \ref{alg:ukfupdatepersensor}, \ref{alg:ut}, and \ref{alg:genmsobservation}, Appendix A)\;
           $(\mu_i^{(p)},\mathbf{P}_{i}^{(p)})\leftarrow$\funcfont{UKFUpdateKP}($k_j,\mu_i^{(p)},\mathbf{P}_{i}^{(p)}\!,c$) (Algorithm \ref{alg:ukfupdatepersensor_kp}, Appendix A)}
    }
    \If{\funcfont{sum}$(A[i,:])>0$}{
        $\mathcal{A}\leftarrow$\funcfont{Append}$(\mathcal{A}, (\ell_i,\mu_i,\mathbf{P}_{i}))$
    }
}
$\mathcal{T}\leftarrow\mathcal{P}_{t-1}\setminus\mathcal{A}$ $\thinspace$\funcfont{/*Tentative Tracks*/}\;

\funcfont{/*Track Initialization \& Termination*/}\;
\For{$c=1:v$}{
    $U[c]\leftarrow$ \funcfont{UnassignedDetection}($A[:,c],B^{(c)}$)\;
}
$G\leftarrow$\funcfont{MeanshiftClustering}($U$)\;
$\mathcal{I}\leftarrow$\funcfont{EmptyList}\quad\funcfont{/* New Tracks */}\;
\For{$l=1:$ \funcfont{length}($G$)}{
    $(\ell_l,\mu_l,\mathbf{P}_{l})\leftarrow$\funcfont{TrackInitialzation}$(G[l])$ \;
    $\mathcal{I}\leftarrow$\funcfont{Append}($\mathcal{I},(\ell_l,\mu_l,\mathbf{P}_{l}))$ \;
}

$\mathcal{T}\leftarrow$\funcfont{DeleteProlongedMissedTracks}$(\mathcal{T})$\;
$(\mathcal{A},\mathcal{I},\mathcal{T})\leftarrow$\funcfont{DeleteDuplicatedTracks}$(\mathcal{A},\mathcal{I},\mathcal{T})$\;
3DPositions\&Poses $\leftarrow$ \funcfont{Extract3DEstimates}$($\funcfont{Append}$(\mathcal{A},\mathcal{I}))$\;
$\mathcal{P}_{t}\leftarrow$\funcfont{Append}$(\mathcal{A},\mathcal{I},\mathcal{T})$\;
\caption{Algorithm iteration at time $t$. \label{alg:sf3dmot}}
\end{algorithm}

First, tracks in the pool are predicted forward using \funcfont{KalmanPrediction}. In the data association step, for each camera $c$ we compute a cost matrix $C$ to assign detections to tracks. If a detection is accepted into the gate of the $i^{th}$ track (determined by the function \funcfont{DetectionGating}, Algorithm \ref{alg:det-gating}, Appendix A), we compute the assignment cost by taking the negative logarithm of the likelihood $q$, which is obtained from the \funcfont{UKFUpdateKS} function (Algorithm \ref{alg:ukfupdatepersensor}, Appendix A). Note that $q$  does not depend on the keypoints. If the cost is greater than $\tau_C$, we set it to infinity. The \funcfont{2DBBoxtoGroundPlane} function (Algorithm \ref{alg:2DtoGP}, Appendix A) transforms the center of the bottom edge of the 2D bounding box into a 3D coordinate using back-projection with homogeneous coordinates. This 3D coordinate represents the ground plane position of the 3D ellipsoid. The \funcfont{2DLinearAssignment} determines a row assignment, represented as a vector of length $n$, where $n$ is the number of tracks in the track pool. This assignment minimizes the cost matrix $C$. The $i^{th}$ component of the row assignment vector is the index of the column that is assigned to the $i^{th}$ row. This value is 0 if there is no column assigned to the row.

The \funcfont{UKFUpdateKS} function is used to update the kinematic and shapes of tracks using the bounding box detections determined by the optimal detection to track assignment from all cameras (outputs of the \funcfont{2DLinearAssignment}). Similarly, the \funcfont{UKFUpdateKP} function (Algorithm \ref{alg:ukfupdatepersensor_kp}, Appendix A) is used to update the keypoint components of the track states. The function \funcfont{UnscentedTransform} (Algorithm \ref{alg:ut}, Appendix A) is implemented based on \cite{van2004sigma}. The function \funcfont{3Dto2DBBox} (Algorithm \ref{alg:genmsobservation}, Appendix A) is an approximation of the quadric projection \cite[pp.~201]{hartley2003multiple} to project a 3D ellipsoid onto the image plane. In this algorithm, we transform six extreme points of the 3D ellipsoid to 2D image plane coordinate, and determine the (axis-aligned) 2D bounding box enclosing all the transformed points, as illustrated in Fig. \ref{fig:ukf_example}, Section \ref{sec:Algorithm}.

After track update, tracks that have at least one detection assigned to them are pushed to the active track list $\mathcal{A}$. Tracks that are in the track pool $\mathcal{P}_{t-1}$ but not appended to the active track list are pushed to the tentative track list $\mathcal{T}$. Unassigned detections are supplied to the \funcfont{MeanshiftClustering} function (based on the mean-shift clustering algorithm for 3D track initialization proposed in \cite{linh2024inffus}) to generate an array of lists $G$. Each element of $G$ is a list of unassigned detections that are clustered together. Each cluster is passed to the \funcfont{TrackInitialization} function to generate a new track, which is then pushed to the list $\mathcal{I}$. The \funcfont{DeleteProlongedMissedTracks} function deletes tracks that are consecutively miss-detected for a number of frames. The \funcfont{DeleteDuplicatedTracks} function deletes the duplicated tracks by computing the pair-wise 3D IoU scores and eliminating tracks with shorter lifespans when the similarity exceeds a threshold. The \funcfont{Extract3DEstimates} function returns the 3D track estimates, which are the mean 3D position, shape, and keypoints of the tracks, extracted from mean states $\mu$ of new ($\mathcal{I}$) and active ($\mathcal{A}$) tracks.

\section{Derivation from the Multi-Sensor GLMB Filter\label{subsec:Relevance-MSGLMB}}

For completeness, this section presents the rationale behind our algorithm which is a drastic
approximation of the multi-sensor generalized labeled multi-Bernoulli (MS-GLMB) filter \cite{vo2019multi} that was adapted
for 3D multi-view multi-object tracking in \cite{ong2020bayesian,linh2024inffus}.
In essence, the MS-GLMB filter propagates the multi-object filtering
density $\boldsymbol{\pi}$ that contains all information about the
set of objects (and their trajectories), given all the multi-sensor detections up to the current
time. The multi-object filtering density is a GLMB, completely characterized
by the parameters, i.e., \cite{vo2019multi}, 
\[
\boldsymbol{\pi}=\left\{ (\omega^{(I,\xi)},p^{(\xi)}):(I,\xi)\in\mathcal{F}_{L}\times\Xi\right\} ,
\]
where: $(I,\xi)$ is a (data association) hypothesis, consisting of
a set $I$ of object IDs and an association map history $\xi=\gamma_{0:t}$
; $\omega^{(I,\xi)}$ is the hypothesis weight; $p^{(\xi)}$ is the
state density of an object with association history $\xi$; and $\mathcal{F}_{L}$,
$\Xi$ denote, respectively, the space of all $I$ and $\xi$. 

Computing the GLMB filtering density at each time requires solving
a series of NP-hard multi-dimensional assignment problems to determine
a prescribed number of highest-weighted hypotheses \cite{vo2019multi}.
Each hypothesis weight $\omega^{(I,\xi)}$ (see equation (13) of \cite{vo2019multi})
depends on the object survival/birth probabilities and the \textit{multi-camera
detection} weight. Specifically, for an object $\ell\in I$, and an
association map $\gamma^{(c)}$ of camera $c$, if 
$\ell$ generates a detection, the \textit{single-camera
detection} weight is given by $\bar{\psi}^{(c,\gamma^{(c)})}(\ell)=P_{D}^{(c)}q^{(\ell,c)}(b_{\gamma^{(c)}(\ell)}^{(c)})$,
where $P_{D}^{(c)}$ is the detection probability for camera $c$,
and $q^{(\ell,c)}(\cdot)$ is defined in equation (\ref{eq:q-per-cam}). If $\ell$ is miss-detected, then $\bar{\psi}^{(c,\gamma^{(c)})}(\ell)=1-P_{D}^{(c)}$. We approximate the multi-camera detection weight by the product of all single-camera detection weights, taken over all cameras.

\begin{figure*}[h!]
\begin{centering}
\includegraphics[width=2\columnwidth]{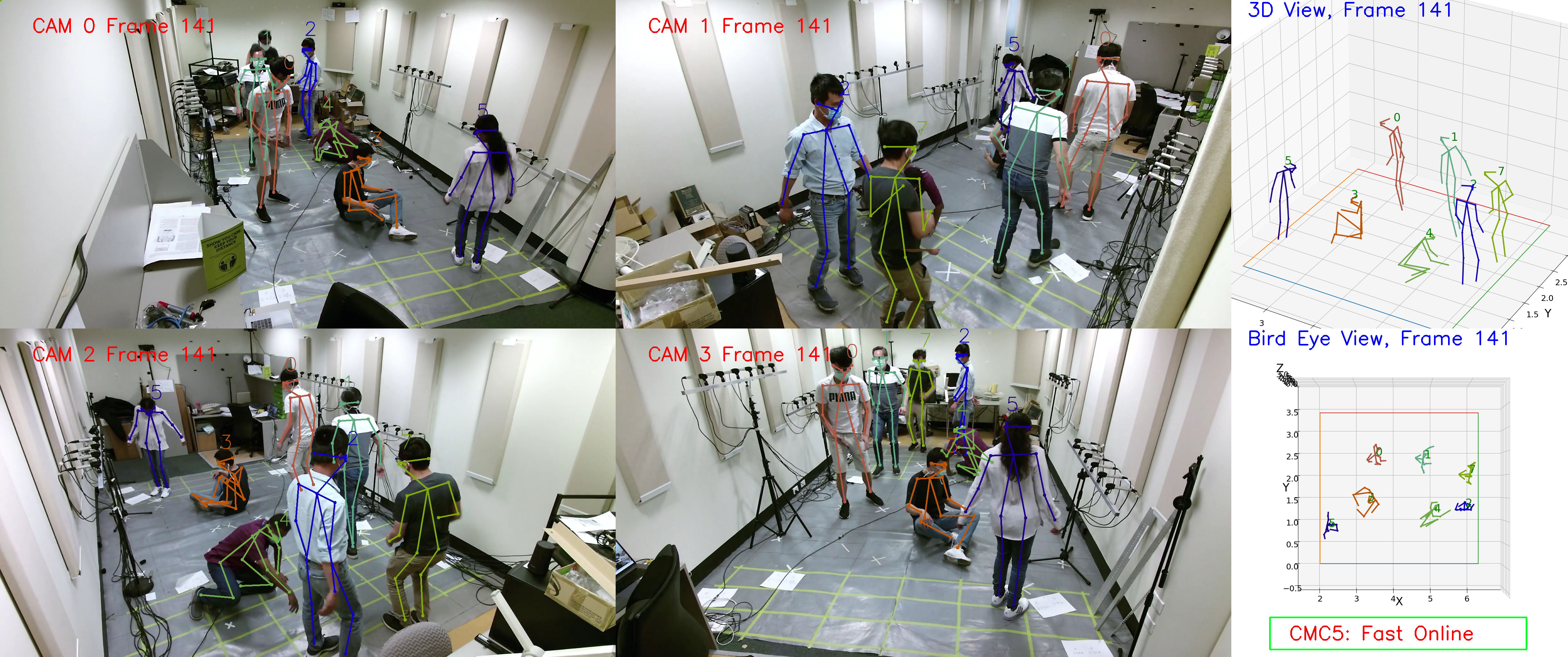}
\includegraphics[width=2\columnwidth]{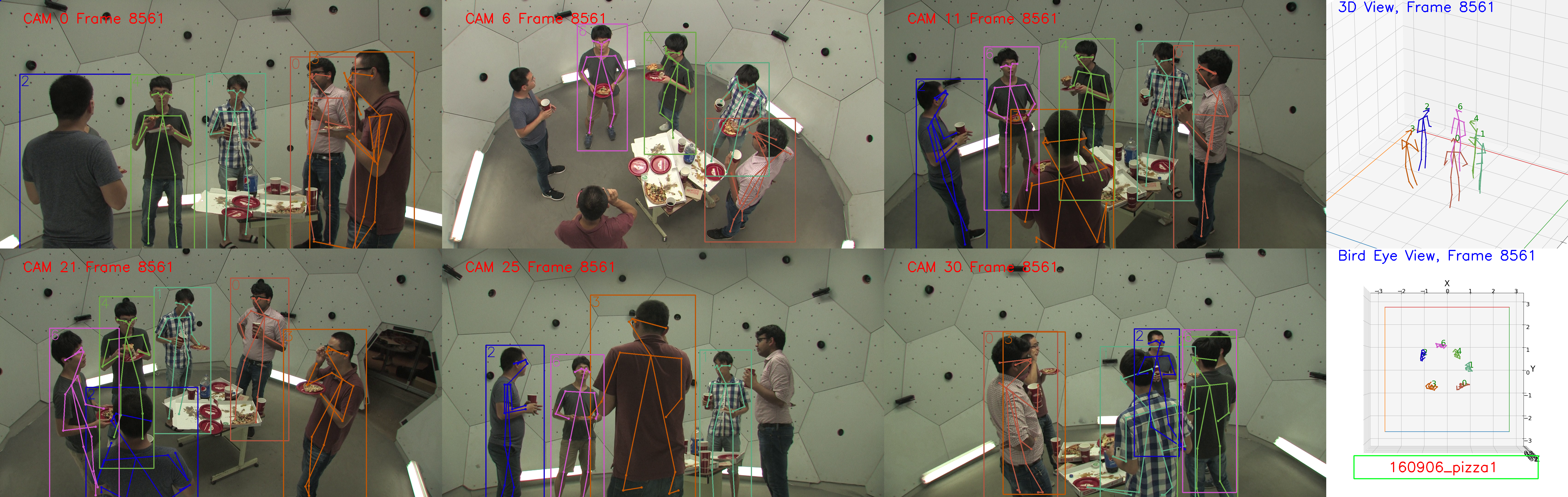}
\par\end{centering}
\caption{Bird's-eye and 3D views of the estimated 3D positions and poses in the CMC (top two rows) and CMU (bottom two rows) datasets. The inputs consist of 2D detections from different cameras. Our algorithm accurately estimates the 3D poses of multiple persons having different poses, i.e., standing, eating, walking, jumping, and falling on the ground. \label{fig:qual_pose}}
\end{figure*}

To reduce the computations, we approximate the GLMB filtering density by its single
most significant hypothesis, given that the track deaths and births
are captured accurately by a separate track initialization/termination
procedure. This is a reasonable assumption due to the high detection
quality (of SOTA 2D image detectors) with relatively low false negative rates. Consequently, we restrict
ourselves to hypotheses $(I,\gamma_{0:t})$, where $I$ is a set of IDs
of existing tracks, and $\gamma_{0:t}$ is the sequence of optimal
association maps from the initial time to $t$. This means the hypothesis
weight (equation (13) of \cite{vo2019multi}) reduces to
\begin{equation}
\omega^{(I,\gamma_{0:t})}\appropto\prod_{c\in\{1:v\}}\prod_{\ell\in I}\bar{\psi}^{(c,\gamma_{t}^{(c)})}(\ell),
\end{equation}
and since the false negative rate is low, i.e., $P_{D}^{(c)}\approx1$,
we have
\begin{equation}
\bar{\psi}^{(c,\gamma_{t}^{(c)})}(\ell)\simeq(1-\delta_{\gamma_{t}^{(c)}(\ell)}[0])q^{(\ell,c)}(b_{\gamma_{t}^{(c)}(\ell)}^{(c)})+\epsilon\delta_{\gamma_{t}^{(c)}(\ell)}[0],
\end{equation}
where $\epsilon$ is a small positive number. Taking the negative
logarithm of the hypothesis weight yields
\begin{equation}
C^{(I)}(\gamma)=\sum_{c=1}^{v}\sum_{\ell\in I}C^{(c)}(\ell,\gamma^{(c)}(\ell)),\label{eq:total-cost}
\end{equation}
where $\gamma=[\gamma^{(1)},...,\gamma^{(v)}]$ and  $C^{(c)}(\ell,\gamma^{(c)}(\ell))$
is defined in equation (\ref{eq:match-cost}) in Section \ref{sec:Algorithm}. Hence, minimizing the cost in equation
(\ref{eq:total-cost}) approximately maximizes the hypothesis weight. Further, since
each camera's association map $\gamma^{(c)}$ contributes independently
to the total cost, it can be minimized by minimizing each
outer term independently, i.e., by solving the optimization problem
in equation (\ref{eq:cost-per-cam}), Section \ref{sec:Algorithm}.

\section{Experimental Results}\label{sec:experiment_sfmcmt}
In this section, we evaluate the performance of our algorithm on multiple multi-camera multi-object tracking datasets, including Curtin Multi-Camera (CMC) \cite{ong2020bayesian}, WILDTRACK (WT) \cite{chavdarova2018wildtrack}, and MultiviewX \cite{hou2020multiview}. We test the performance in standard configuration when all cameras are in operation and multi-camera re-configuration when cameras are on/off at different instances, simulating camera failures. Further, we also evaluate the performance of our method on 3D pose estimation datasets including Campus, Shelf \cite{belagiannis20143d}, and CMU Panoptic (CMU) \cite{joo2015panoptic}. In Fig. \ref{fig:qual_pose}, we show examples of our algorithm output at one time step for CMC and CMU datasets.

\subsection{Standard Multi-Camera Configuration}\label{subsec:exp-standard-tracking}
\begin{table*}[h!]
\centering{}
\global\long\def\arraystretch{1.2}%
 \caption{Tracking performance of different algorithms on CMC, WT, and MultiviewX datasets. The best result for each dataset
is \underline{\textbf{bolded and underlined}}. Standard deviation over Monte Carlo runs for probabilistic algorithms is reported in parentheses. FPS$^{\text{*}}$ denotes the processing speed excluding 2D detectors run-time, and FPS denotes the end-to-end processing speed.  \label{tbl:standard-performance}}
 \scriptsize
\begin{tabular}{|c|c|cccccccc|}
\hline 
Sequence & Tracker & FP$\downarrow$ & FN$\downarrow$ & IDs$\downarrow$ & MOTA$\uparrow$ & IDF1$\uparrow$ & OSPA\!$^{\text{(2)}}$\!$\downarrow$ & FPS$^{\text{*}}$$\uparrow$ & FPS$\uparrow$\tabularnewline
\hline 
\hline 
\multirow{4}{*}{CMC1} & Ours & 0 & \underline{\textbf{3}} & \underline{\textbf{0}} & \underline{\textbf{99.5}} & \underline{\textbf{99.8}} & \underline{\textbf{0.30}} & \underline{\textbf{4011.70}} & \underline{\textbf{6.36}}\tabularnewline
 & MV-GLMB-AB \cite{linh2024inffus} & 0	& 5 & 0 & 99.2(0.0) & 99.6(0.0) & 0.3(0.0) & 11.69 & 4.12\tabularnewline
 & MV-GLMB \cite{ong2020bayesian} & 55 & 87 & 3 & 77.7(5.45) & 50.4(1.64) & 0.91(0.01) & 0.65 & 0.59\tabularnewline
 & MS-GLMB \cite{vo2019multi} & 13 & 87 & 3 & 84.1(0.65) & 52.4(0.29) & 0.89(0.01) & 0.75 & 0.67\tabularnewline
\hline 
\multirow{4}{*}{CMC2} & Ours & \underline{\textbf{2}} & \underline{\textbf{22}} & \underline{\textbf{0}} & \underline{\textbf{98.8}} & \underline{\textbf{99.4}} & \underline{\textbf{0.36}} & \underline{\textbf{918.15}} & \underline{\textbf{6.36}}\tabularnewline
 & MV-GLMB-AB \cite{linh2024inffus} & 0 & 34 & 1 & 98.3(0.07) & 94.7(4.79) & 0.39(0.02) &  8.29 & 3.61\tabularnewline
 & MV-GLMB \cite{ong2020bayesian}  & 431 & 334 & 80 & 59.3(2.06) & 33.7(1.58) & 0.94(0.01) & 0.08 & 0.08\tabularnewline
 & MS-GLMB \cite{vo2019multi}  & 214 & 454 & 72 & 64.3(5.27) & 36.7(2.71) & 0.91(0.01) & 0.02 & 0.02\tabularnewline
\hline 
\multirow{4}{*}{CMC3} & Ours & \underline{\textbf{11}} & \underline{\textbf{53}} & \underline{\textbf{0}} & \underline{\textbf{97.7}} & \underline{\textbf{98.9}} & \underline{\textbf{0.32}} & \underline{\textbf{564.26}} & \underline{\textbf{6.31}}\tabularnewline
 & MV-GLMB-AB \cite{linh2024inffus} & 29 & 63 & 6 & 96.5(0.23) & 92.5(4.43) & 0.36(0.03) & 5.30 & 2.9\tabularnewline
 & MV-GLMB \cite{ong2020bayesian}  & 608 & 475 & 121 & 57.4(2.72) & 36.7(2.53) & 0.93(0.01) & 0.04 & 0.04\tabularnewline
 & MS-GLMB \cite{vo2019multi}  & 269 & 625 & 122 & 64.0(3.12) & 39.1(2.22) & 0.91(0.01) & 0.02 & 0.02\tabularnewline
\hline 
\multirow{4}{*}{CMC4} & Ours  & \underline{\textbf{0}} & \underline{\textbf{8}} & \underline{\textbf{0}} & \underline{\textbf{98.0}} & \underline{\textbf{99.0}} & \underline{\textbf{0.25}} & \underline{\textbf{439.03}} & \underline{\textbf{6.30}}\tabularnewline
 & MV-GLMB-AB \cite{linh2024inffus} & 1 & 19 & 0 & 94.7(0.25) & 97.3(0.16) & 0.26(0.0) &  5.68 & 3.01\tabularnewline
 & MV-GLMB \cite{ong2020bayesian}  & 28 & 231 & 4 & 34.6(11.27) & 47.9(9.87) & 0.86(0.03) & 0.07 & 0.07\tabularnewline
 & MS-GLMB \cite{vo2019multi}  & 36 & 207 & 5 & 38.5(9.51) & 53.5(5.37) & 0.84(0.02) & 0.05 & 0.05\tabularnewline
\hline 
\multirow{4}{*}{CMC5} & Ours &  \underline{\textbf{62}} & \underline{\textbf{285}} & \underline{\textbf{30}} & \underline{\textbf{89.9}} & \underline{\textbf{56.3}} & \underline{\textbf{0.91}} & \underline{\textbf{195.05}} & \underline{\textbf{6.22}}\tabularnewline
 & MV-GLMB-AB \cite{linh2024inffus} & 96 & 382 & 39 & 86.1(0.62) & 44.4(3.21) & 0.93(0.01) & 3.75 & 2.37\tabularnewline
 & MV-GLMB \cite{ong2020bayesian} & 464 & 859 & 102 & 61.6(5.89) & 22.7(2.87) & 0.95(0.01) & 0.06 & 0.06\tabularnewline
 & MS-GLMB \cite{vo2019multi}  & 491 & 887 & 131 & 59.3(4.98) & 22.7(3.09) & 0.96(0.01) & 0.07 & 0.07\tabularnewline
\hline 
\multirow{4}{*}{WT} & Ours & 4286 & \underline{\textbf{590}} & \underline{\textbf{112}} & \underline{\textbf{47.6}} & \underline{\textbf{75.0}} & 0.76 & \underline{\textbf{95.67}} & \underline{\textbf{5.99}}\tabularnewline
 & MV-GLMB-AB \cite{linh2024inffus} & 3664 & 1996 & 193 & 38.5(2.76) & 63.1(2.36) & \underline{\textbf{0.72(0.03)}} &  0.12 & 0.12\tabularnewline
 & MV-GLMB \cite{ong2020bayesian} & \underline{\textbf{2286}} & 3619 & 310 & 34.7(1.83) & 47.4(1.55) & 0.79(0.01) & 0.02 & 0.02\tabularnewline
 & MS-GLMB \cite{vo2019multi} &  3619 & 4726 & 281 & 9.4(8.5) & 34.2(7.36) & 0.87(0.03) & 0.02 & 0.02\tabularnewline
 \hline 
\multirow{4}{*}{MultiviewX} & Ours & 2261 & \underline{\textbf{677}} & \underline{\textbf{477}} & \underline{\textbf{78.0}} & \underline{\textbf{72.1}} & 0.74(0.0) & \underline{\textbf{98.70}} & \underline{\textbf{6.03}}\tabularnewline
 & MV-GLMB-AB \cite{linh2024inffus} & \underline{\textbf{1753}} & 2034 & 460 & 72.6(1.4) & 55.8(1.4) & \underline{\textbf{0.64(0.01)}} & 0.12 & 0.12\tabularnewline
 & MV-GLMB \cite{ong2020bayesian} & 11443 & 2505 & 1337 & 1.3(3.15) & 42.1(1.03) & 0.83(0.01) & 0.02 & 0.02\tabularnewline
 & MS-GLMB \cite{vo2019multi} &  11501 & 2520 & 1328 & 0.9(3.55) & 41.9(0.98) & 0.83(0.01) & 0.02 & 0.02\tabularnewline
\hline 
\end{tabular}
\end{table*}

\begin{figure}[h!]
\begin{centering}
\includegraphics[width=0.9\columnwidth]{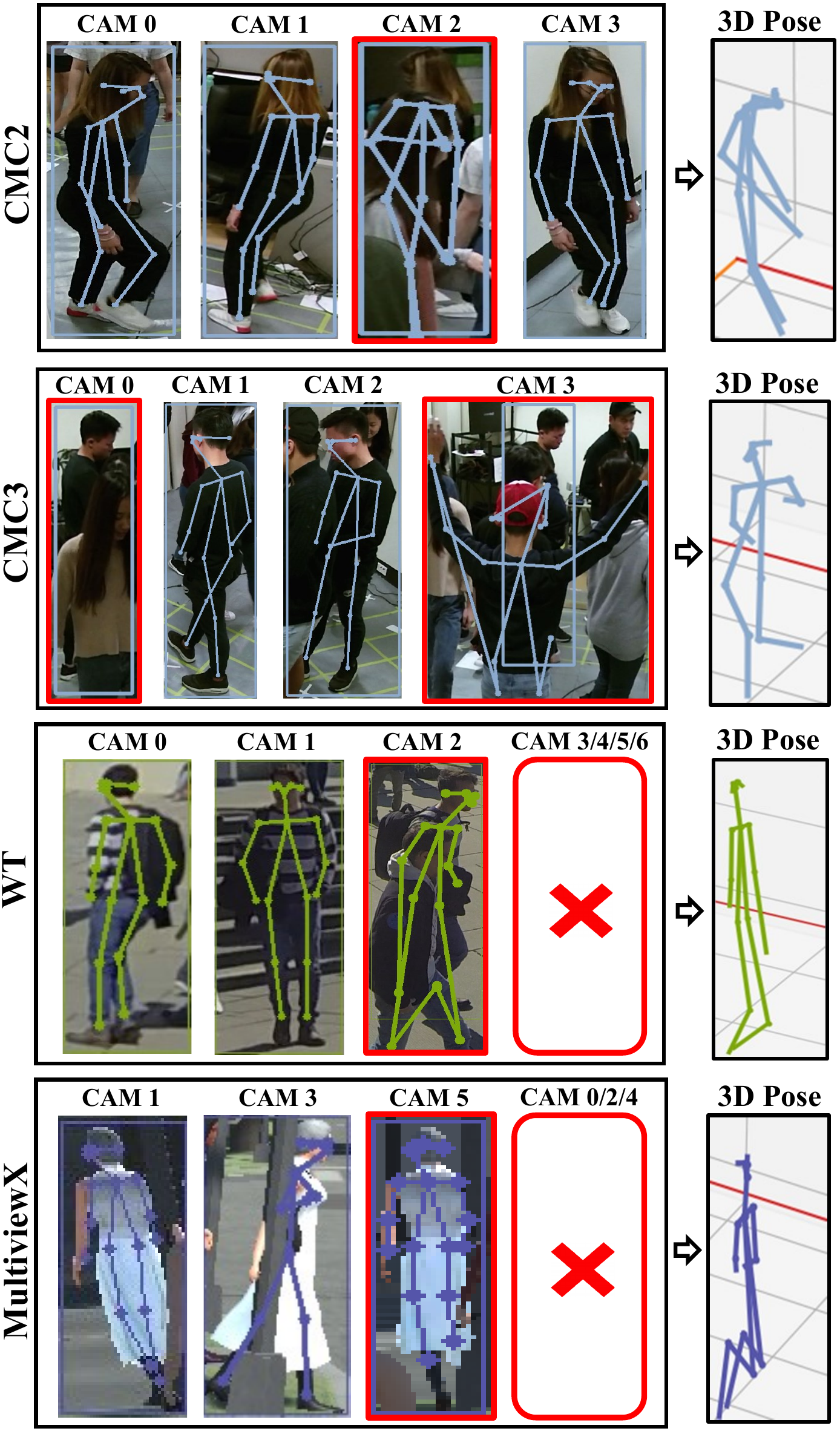}
\par\end{centering}
\caption{Typical incorrect 3D pose estimates from our algorithm in CMC, WT, and MultiviewX datasets. Red rectangles with ``x" mark denote miss-detection. Red ``x" mark denotes AlphaPose is unable to detect or produces incorrect 2D poses. Severe miss-detection and inaccurate 2D detection on multiple cameras lead to the inaccuracy of our 3D pose estimation.\label{fig:bad_pose}}
\end{figure}

In this subsection, we consider the
CMC dataset consisting
of 5 sequences recorded with 4 cameras, WT
dataset  consisting of one sequence
recorded with 7 cameras, and MultiviewX synthetic dataset consisting of 6 cameras. Different sequences in the CMC dataset
have different levels of difficulty. In particular, sequences 4 and 5 of the CMC dataset
have people constantly jumping and falling, posing significant difficulty for 
 3D pose estimation algorithms. Note that while the CMC dataset has the ground
truth of real-world 3D position of objects (including their size),
the WT and MultiviewX datasets only have the ground truth of object feet location. Notably, we do not use any portion of the datasets for training the 2D detectors, and evaluation is performed on the entire datasets.

To evaluate 3D tracking performance, we use various performance measures to benchmark the algorithm. We use the CLEARMOT \cite{bernardin2008evaluating} measure with multi-object tracking accuracy (MOTA)
as the main score and other component scores including numbers of
false positive (FP), false negative (FN) and ID
switches (IDS). We also report the IDF1 score \cite{ristani2016performance}, and use the OSPA\!$^{\text{(2)}}$ metric
\cite{beard2020asolution,rezatofighi2020trustworthy} to measure the
tracking error. For CMC dataset, we use the 3D GIoU distance \cite{rezatofighi2019generalized} normalized
between 0 and 1 as the pair-wise distance between objects, while for
the WT and MultiviewX datasets, we use Euclidean distance between object feet locations
as pair-wise distance. For CLEARMOT and track identity measure, the
pair-wise distance threshold to reject/accept a match is 0.5 for GIoU
distance and 1 meter for Euclidean distance. For OSPA\!$^{\text{(2)}}$ metric, the
cut-off parameter is set to 1 for GIoU distance and 1 meter for Euclidean
distance.  Since all tested datasets in this subsection do not have ground truth keypoints, we provide videos showing 3D pose estimation in the supplementary materials for qualitative evaluation. Additional evaluation with popular 3D keypoint estimation datasets will be provided in Subsection \ref{subsec:3dpose-exp}.


We use YOLOX \cite{ge2021yolox} algorithm to generate 2D bounding box detection and AlphaPose \cite{fang2022alphapose} to estimate 2D pose in each detected bounding box. The algorithms were tested on a desktop equipped with an Intel(R) Core(TM) i7-8700 CPU @ 3.20GHz. The 2D bounding box and pose detection tasks were accelerated using an NVIDIA GeForce GTX 1080 Ti GPU, achieving average speeds of 7 frames per second (FPS) and 70 FPS, respectively. In contrast, no GPU acceleration was used for the 3D tracking task. To assess the efficiency, we report the processing speed excluding the 2D detection run-time (denoted as FPS$^{\text{*}}$), and the end-to-end FPS rate.
We compare our tracking algorithm
with the  MV-GLMB-AB \cite{linh2024inffus}, MV-GLMB \cite{ong2020bayesian} and MS-GLMB \cite{vo2019multi}
filters. Since these filters are probabilistic, we report their
mean performance over 25 Monte Carlo runs.  

Observe Table \ref{tbl:standard-performance}, across all sequences, our method outperforms MV-GLMB and MS-GLMB significantly and exhibits similar performance to the MV-GLMB-AV filter
in terms of tracking accuracy. In particular, the algorithm performs well in CMC1-4 and MultiviewX due
to the accurate detections of those sequences. However, the tracking
accuracy drops in sequences CMC5 and WT due to the higher rate of object miss-detection compared to the other sequences. The FPS$^{\text{*}}$ and FPS columns of Table \ref{tbl:standard-performance} show that
our method is much faster than the others which demonstrate
its real-time 3D tracking and pose estimation potential. Further, they show that the contribution of the proposed tracking module to the overall processing time is marginal compared to the 2D detectors.

The significant performance gaps, especially on the IDF1 score and
OSPA\!$^{\text{(2)}}$ error, are because the MS-GLMB and MV-GLMB filters generate a high
number of false tracks due to the fixed birth track initialization process.
When there are more false tracks, ID switches also have a higher tendency
to occur. Ours and MV-GLMB-AB filter have an improved track initialization process based on mean-shift clustering, hence better tracking accuracy. Moreover, the low number of ID switches in our method is also due to the fact that we do not discard a track immediately after a miss-detection; instead, we store it as a tentative track and recall it later when the object is re-detected.  Further, MV-GLMB-AB, MV-GLMB and MS-GLMB filters use
Gibbs sampler to perform data association. Although the Gibbs sampler is able to select diverse hypotheses to represent the GLMB density, it might miss the optimal solution due to its random nature. Moreover, the MV-GLMB
filter incorporates an occlusion model to resolve object occlusion.
However, we observe that the model assigns low detection
probability to the objects even though they are only partially
occluded. Indeed, these objects are detected relatively well by the YOLOX detector. Hence,  the
mismatch between the model and the observed data also contributes to performance degradation.

In addition to the MS-GLMB-based filters, we also evaluate our algorithm against other deep-learning-based 3D trackers and classical trackers using 3D-training-based detectors on WT dataset. The results, reported in Appendix B, show that our method’s performance is comparable to methods trained on 10\%–30\% of the WT dataset (and evaluated on the remainder), but it is outperformed by methods trained on over 50\% of the dataset. The inferior performance of our method stems from the lower quality of the 2D detections compared to the 3D detections produced by trained 3D detectors, as the YOLOX detector was not trained on any portion of the WT dataset.

Among the techniques included in this study, only our method can estimate 3D poses. Fig. \ref{fig:qual_pose} shows examples of the outputs from our algorithm. We include the videos of our estimation results on CMC3, CMC5 (the most crowded sequences in the CMC dataset), WT, and MultiviewX sequences in the supplementary materials. We observe that our algorithm performs generally well across all sequences, especially when the object density is not high. However, when some 2D keypoints are miss-detected or incorrect, the estimated 3D pose becomes inaccurate. The errors occur more often in crowded scenes, such as ones in WT and CMC5 sequences. In Fig. \ref{fig:bad_pose}, we show our algorithm's typical incorrect 3D pose estimation.

\subsection{Multi-Camera Reconfiguration\label{sec:multi-cam-reconf_sfmcmt}}
\begin{figure}[h!]
\begin{centering}
\includegraphics[width=0.9\columnwidth]{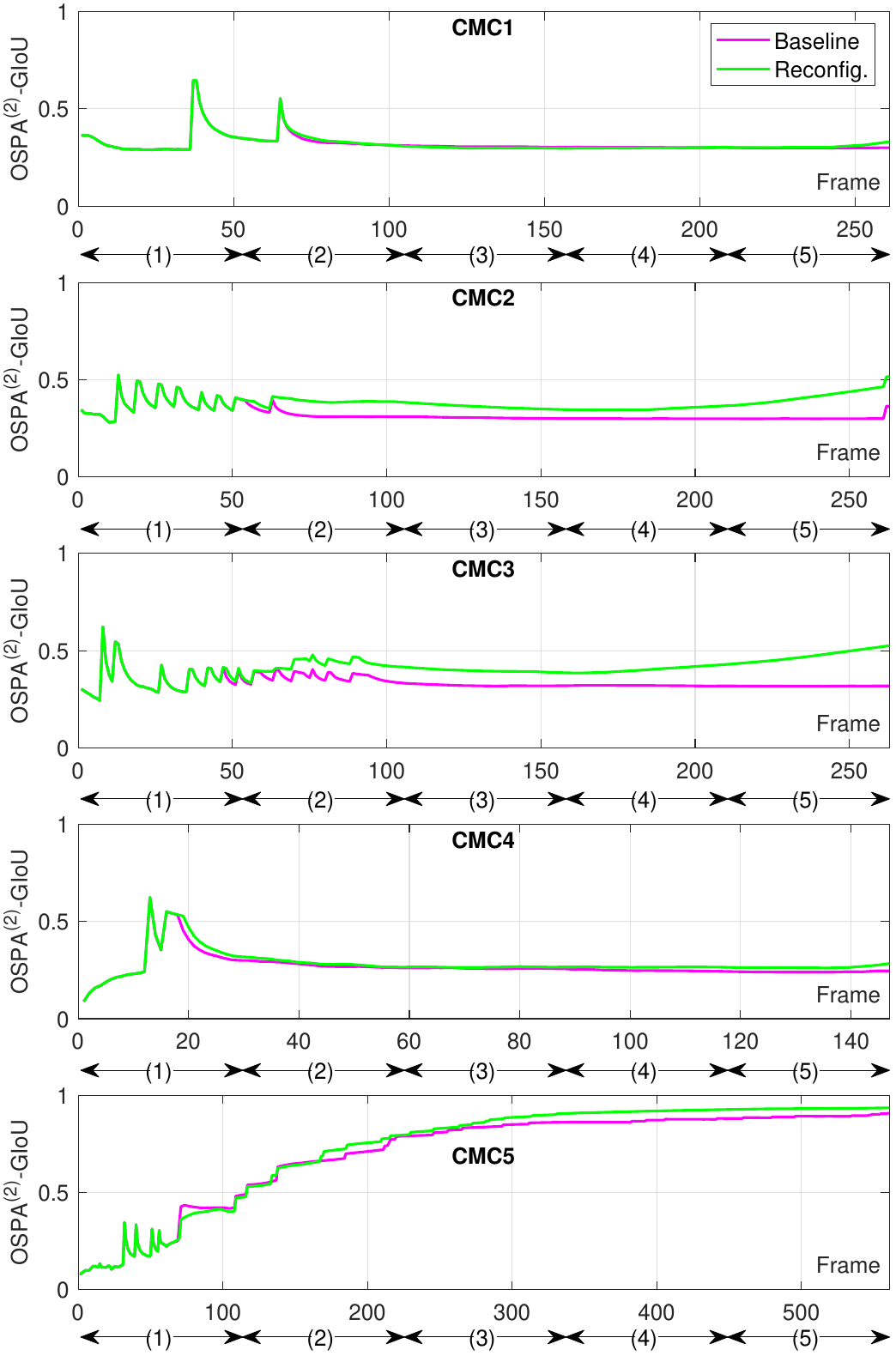}
\par\end{centering}
\caption{Camera reconfiguration setting and the OSPA$^{(2)}$ error of the
proposed algorithm with YOLOX detector. Configuration (1): all cameras
at positions 1, 2, 3, and 4 are on. Configuration (2): three cameras
on at positions 2, 3, and 4. Configuration (3): three cameras on at
random positions. Configuration (4): two cameras on at positions 1
and 3. Configuration (5): two cameras on at positions 2 and 4.\label{fig:configcam_sfmcmt}}
\end{figure}
To test the robustness of the algorithm in the scenarios of camera failures, in this experiment, we occasionally disconnect
the cameras and reconnect them again to demonstrate the algorithm
capability in tracking objects when the camera configuration
changes on-the-fly in CMC dataset. The configurations (on/off time) of the camera
are depicted in Fig. \ref{fig:configcam_sfmcmt}. In this same figure, we plot the OSPA\!$^{\text{(2)}}$
error per time step for both the standard (baseline, all cameras are always on) and the reconfiguration
settings. Note that the OSPA\!$^{\text{(2)}}$ error at time $t$ is the OSPA\!$^{\text{(2)}}$
error evaluated from the initial time step up to $t$. We observe that tracking performance in the reconfiguration settings is similar 
to the baseline in CMC1 and CMC4 sequences. This is because objects are well-separated in those two scenarios. Thus, turning the
cameras off does not significantly reduce the detection quality (compared to the baseline configuration), hence the similarity in tracking accuracy to the baseline. 
In other sequences, which have higher object density, the tracking performance slightly degrades after the cameras
are turned off since  some objects are miss-detected
and incorrectly terminated. We observe that the trends of pose estimation performance follow the trends of the tracking performance in this experiment.

\subsection{Performance Evaluation on 3D Keypoint Datasets}\label{subsec:3dpose-exp}

In this subsection, we evaluate the performance of our technique on popular datasets for 3D keypoint estimation, including the Campus, Shelf \cite{belagiannis20143d}, and CMU PanOptic (CMU) \cite{joo2015panoptic}. Campus is a dataset consisting of three people interacting with each other in an outdoor environment, captured with three calibrated cameras. Compared with Campus, Shelf dataset is more complex, which consists of four people disassembling a shelf at a close range. CMU dataset is recorded in a studio equipped with a large number of cameras and features multiple individuals engaged in social activities. To showcase the efficiency, we report both the 3D processing speed, e.g., our 3D tracker excluding 2D detection run-time, denoted as FPS$^\text{*}$, and the overall FPS rate. The evaluation was done using the same desktop computer as one used in Subsection \ref{subsec:exp-standard-tracking}. For a fair comparison, we only report processing speeds for methods for which we can obtain codes to test on our hardware. Note that we only use GPU-acceleration for the 2D detectors but not for the proposed 3D tracking module.
\begin{table*}[h!]
\centering{}
 \caption{PCK scores for various methods on Campus dataset frames 350 to 470 and 650 to 750, and Shelf dataset frames 300 to 600 (following the datasets guidelines for evaluation). The highest value (best result) for each column
 is \underline{\textbf{bolded and underlined}}.\label{tbl:posecompare}} 
\centering{}
 \scriptsize
\setlength{\tabcolsep}{1.8mm}{\begin{tabular}{|l|c|cccccc|cccccc|}
\cline{3-14}
 \multicolumn{1}{c}{ } &  & \multicolumn{6}{c|}{Campus Dataset} & \multicolumn{6}{c|}{Shelf Dataset} \tabularnewline
\hline
Method & 3D Training Free & Actor 1 & Actor 2 & Actor 3 & Average & FPS$^{\text{*}}$ & FPS &  Actor 1 & Actor 2 & Actor 3 & Average & FPS$^{\text{*}}$ & FPS\tabularnewline
\hline
3DPSv1 \cite{belagiannis20143d} & \xmark & 82.0 & 72.4 & 73.7 & 75.8 & - & - & 66.1 & 65.0 & 83.2 & 71.4 & - & -\tabularnewline
\rowcolor{mygray}
3DPSv2 \cite{belagiannis2015multiple} & \xmark & 83.0 & 73.0 & 78.0 & 78.0 & - & - & 75.0 & 67.0 & 86.0 & 76.0 & - & -\tabularnewline
3DPSv3 \cite{belagiannis20153d} & \xmark & 93.5 & 75.7 & 84.4 & 84.5 & - & - & 75.3 & 69.7 & 87.6 & 77.5 & - & -\tabularnewline
\rowcolor{mygray}
CRF \cite{ershadi2018multiple} & \xmark & 94.2 & 92.9 & 84.6 & 90.6 & - & - & 93.3 & 75.9 & 94.8 & 88.0 & - & -\tabularnewline
MVPose \cite{dong2019fast} & \xmark & \underline{\textbf{97.6}} & \underline{\textbf{93.3}} & \underline{\textbf{98.0}} & \underline{\textbf{96.3}} & 11.80 & 0.13 & \underline{\textbf{98.8}} & \underline{\textbf{94.1}} & \underline{\textbf{97.8}} & \underline{\textbf{96.9}} & 6.93 & 0.41\tabularnewline
\rowcolor{mygray}
MVGFormer \cite{liao2024multiple} & \xmark & 48.0 & 62.3 & 73.3 & 61.2 & 1.9 & 1.8 & 89.8 & 85.7 & 88.0 & 87.8 & 3.0 & 2.81\tabularnewline
Ours & \cmark & 95.5 & 83.1 & 81.8 & 86.8 & \underline{\textbf{1829.72}} & \underline{\textbf{9.95}} & 90.4 & 66.0 & 93.9 & 83.4 & \underline{\textbf{606.63}} & \underline{\textbf{9.81}}\tabularnewline 
\hline
\end{tabular}}
\end{table*}

On Campus and Shelf datasets, we compare our algorithm with state-of-the-art techniques in the literature including: 3D Pictorial Structures (3DPS)- 3DPSv1 \cite{belagiannis20143d}, 3DPSv2 \cite{belagiannis2015multiple}, 3DPSv3 \cite{belagiannis20153d}; Conditional Random Field (CRF) \cite{ershadi2018multiple}; MVPose \cite{dong2019fast}; and MVGFormer \cite{liao2024multiple}. We note that the MVGFormer model is only trained on CMU dataset but not on Campus or Shelf dataset, and the 2D detectors used in our method are not trained on either Shelf, Campus, or CMU dataset. 
\begin{figure}[h!]
\begin{centering}
\includegraphics[width=0.9\columnwidth]{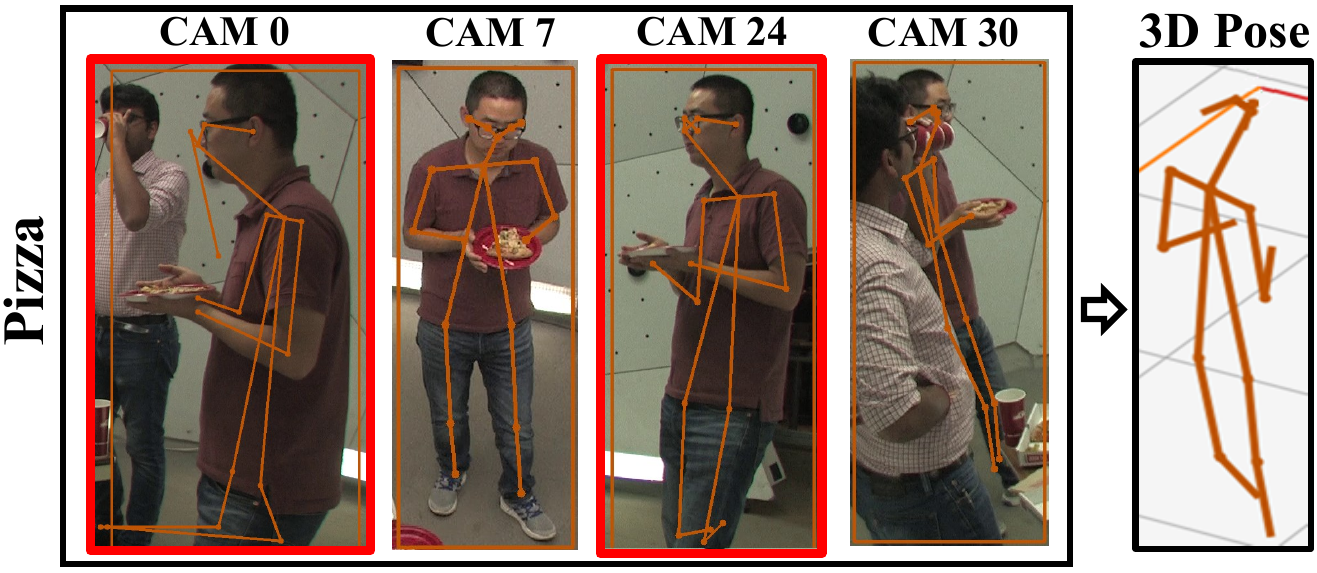}
\par\end{centering}
\caption{Typical incorrect 3D pose estimate from our algorithm in CMU dataset. We only show 2D detection from 4 out of 31 cameras. Red rectangles denote incorrect 2D poses. Note that the person's 2D pose is either miss-detected or incorrectly detected on 22 cameras.\label{fig:bad_posecmu}}
\end{figure}

In Table \ref{tbl:posecompare}, we report the percentage of correct keypoint (PCK) scores for each actor and the average score across all actors, and the run-time in terms of FPS rate on Campus and Shelf datasets. We observe that our method outperforms the 3DPS algorithms on both datasets. Additionally, our approach is comparable to the CRF algorithm, though it performs slightly worse than MVPose. Conversely, our algorithm demonstrates comparable performance with MVGFormer on Shelf dataset. Nonetheless, the MVGFormer algorithm performs significantly worse than ours on Campus dataset, which is due to the mismatch between training and test data in terms of the scene area and level of occlusion. This demonstrates the improved generalizability of our algorithm compared to MVGFormer, as ours was not trained on any of the 3D pose estimation datasets. Meanwhile, with the same hardware setting, our method is significantly faster than MVPose and MVGFormer on both datasets. The notable efficiency of our method stems from the fact that it only needs to process the tracks and solve for a series of 2D linear assignment problems.

\begin{table*}
    \centering
    \caption{MPJPE (measured in mm) for various methods on CMU dataset. The processing speeds are averaged across all sequences. \\The best result for each sequence is \underline{\textbf{bolded and underlined}}.}
    \scriptsize
    \setlength{\tabcolsep}{2mm}{
    \begin{tabular}{l|c|cccccccc} 
    \hline 
    Method & 3D Training Free & Haggling$\downarrow$ &  Ian$\downarrow$ & Band$\downarrow$ & Pizza$\downarrow$ & Mean$\downarrow$  & FPS$^{\text{*}}$$\uparrow$ & FPS$\uparrow$\\
    \hline
    DAS \cite{wang2022distribution} & \xmark & {47.8} & 100.7 & {143.9} & {59.5} & {88.0}  & 9.80 & \underline{\textbf{7.33}}\\ 
    \rowcolor{mygray}
    PlaneSweepPose \cite{lin2021multi} & \xmark & 15.5 & \underline{\textbf{15.2}} & 15.9 & 19.3 & \underline{\textbf{16.5}}  & 0.29 & 0.11\\
    Faster-VoxelPose \cite{ye2022faster} & \xmark & {51.2} & {52.6} & {64.2} & {65.4} & {58.4}  & 0.46 & 0.27\\
    \rowcolor{mygray}
    MVGFormer \cite{liao2024multiple} & \xmark & \underline{\textbf{14.0}} & 27.4 & \underline{\textbf{13.5}} & \underline{\textbf{18.0}} & 18.2  & 9.39 & 5.60\\
    SelfPose3d \cite{srivastav2024selfpose3d} & \xmark & 23.8 & 25.7 & 24.7 & 24.6 & 24.7  & 0.10 & 0.10\\
    \rowcolor{mygray}
    Ours & \cmark & 34.5 & 37.3 & 26.5 & 37.8 & 34.0  & \underline{\textbf{120.52}} & 6.08\\
    \hline
    \end{tabular}}
    \label{tab:sota_panoptic}
\end{table*}


On CMU dataset, we test the performance of our algorithm against DAS \cite{wang2022distribution}, Faster-VoxelPose \cite{ye2022faster}, PlaneSweepPose \cite{lin2021multi}, SelfPose3d \cite{srivastav2024selfpose3d} and MVGFormer \cite{liao2024multiple} on 160422\_haggling1 (Hangling), 160906\_ian5 (Ian),  160906\_band4 (Band), and 160906\_pizza1 (Pizza) sequences, following the evaluation in \cite{srivastav2024selfpose3d}, \cite{ye2022faster} and \cite{liao2024multiple}. We report the mean per joint position error (MPJPE)--measured in millimeters (mm)--in Table~\ref{tab:sota_panoptic}\footnote{All methods are evaluated using ground truth data obtained from http://domedb.perception.cs.cmu.edu/.}. On average, our method outperforms DAS and Faster-VoxelPose in terms of accuracy but underperforms relative to PlaneSweepPose, MVGFormer and SelfPose3d, achieving a mean MPJPE of 34mm. Meanwhile, it is faster than all other methods except DAS, with speeds of 120.5 FPS (excluding 2D detection runtime) and 6.08 FPS for the entire process, although DAS exhibits substantially lower accuracy than ours. An example of the estimation result is previously shown in Fig. \ref{fig:qual_pose} (bottom two rows). Qualitatively, we observe that the pose estimation performance of our algorithm on this dataset is relatively accurate. Although a person might be occluded on multiple cameras, the extensive camera setup capturing multiple views allows for accurate estimation and tracking of multiple individuals' poses. However, if miss-detections or inaccuracies in the detected 2D poses are significant, the performance of the 3D pose estimation may decline, as illustrated in Fig. \ref{fig:bad_posecmu}. This behavior was also observed on the other datasets. A video showcasing the estimation results from our algorithm for this dataset is given in the supplementary materials.

\subsection{Ablation Studies\label{subsec:exp-ablation_sfmcmt}}

\subsubsection{Simulated Miss-Detection Scenarios}
\begin{table}
    \centering
    \caption{MPJPE for our method on CMU dataset under various simulated miss-detection levels. Standard deviations over 25 Monte Carlo simulations are reported in parentheses.}
    \scriptsize
    \setlength{\tabcolsep}{2mm}{
    \begin{tabular}{l|cccc} 
    \hline 
    Sequence Name & Original$\downarrow$ & 20\% Miss$\downarrow$ & 30\% Miss$\downarrow$ & 50\% Miss$\downarrow$\\
    \hline
    Haggling & 34.5 & 38.0 (0.7) & 39.1 (0.5) & 40.3 (0.6)\\
    \rowcolor{mygray}
    Ian & 37.3 & 38.2 (0.3) & 38.3 (0.3) & 39.4 (0.4)\\
    Band & 26.5 & 27.6 (1.9) & 28.2 (1.4) & 29.3 (1.1)\\
    \rowcolor{mygray}
    Pizza & 37.8 & 39.2 (0.8) & 40.2 (0.8) & 41.1 (0.7)\\

    \hline
    \end{tabular}}
    \label{tab:panoptic_abl_conf}
\end{table}

To evaluate the robustness of our method against miss-detection, we conducted an ablation study on the CMU dataset, in which we randomly delete 20\%, 30\% and 50\% of the original detections from each camera to simulate miss-detection events. The experiment was conducted over 25 Monte Carlo runs, with the resulting MPJPE values reported in Table~\ref{tab:panoptic_abl_conf}. The results show that as the false negative rate increased, performance degraded gracefully, indicating that our method is resilient to moderate miss-detection and maintains consistent performance under challenging conditions. For instance, in the Pizza sequence, the MPJPE increased only from 37.8mm to 39.2mm when 20\% of the detections were removed, and slightly to 41.1mm when 50\% of the original detections were deleted. The robust performance stems from our optimal assignment and detection gating procedures, which ensure that correct detections are assigned to the tracks in most cases. Further, when a track is miss-detected by all cameras, it is stored as a tentative track. Upon re-detection by at least one camera, it becomes active again and is included in the estimates, hence preventing the tracker from losing tracks during miss-detection events.

\subsubsection{Effect of Varying Assignment Cost Threshold on Tracking Performance}
\begin{figure}[h!]
\begin{centering}
\includegraphics[width=0.92\columnwidth]{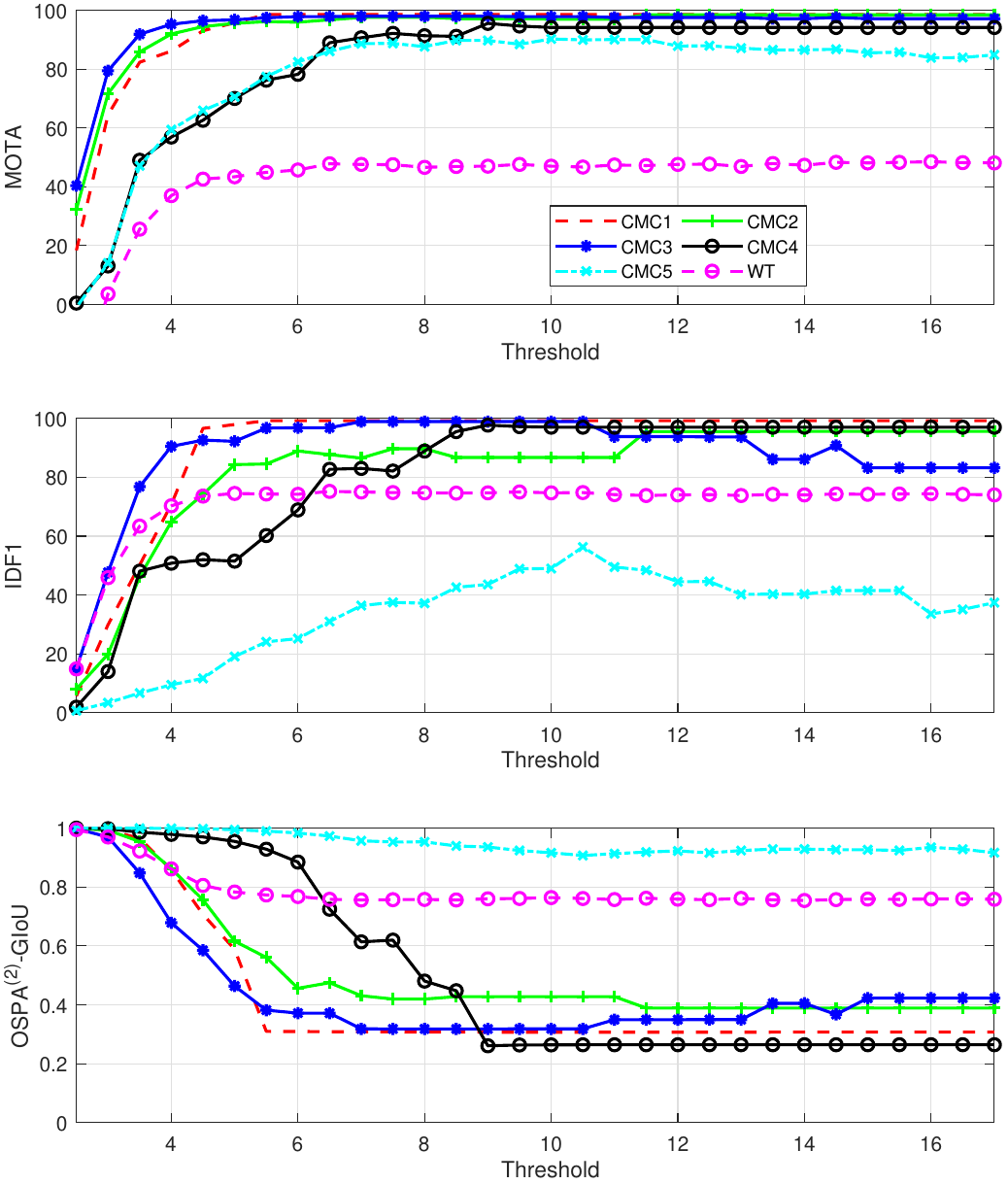}
\par\end{centering}
\caption{Tracking performance with different assignment cost thresholds $\tau_{C}$ in different sequences.\label{fig:la_thresold}}
\end{figure}
We experimentally observe that changing
the assignment cost threshold $\tau_{C}$ could affect
the algorithm's performance, especially if there are miss-detections and false positive detections. If $\tau_{C}$ is low, the optimal assignment
algorithm favors miss-detection assignment. If $\tau_{C}$ is
high, the optimal assignment algorithm might assign false positive detections
to miss-detected tracks or assign detections to false tracks. To illustrate the influence of $\tau_{C}$ on performance, in Fig. \ref{fig:la_thresold}, we
plot the threshold $\tau_{C}$ versus MOTA, IDF1 scores and OSPA\!$^{\text{(2)}}$ error in different sequences. We vary the
threshold from 2 to 17 and observe that when the threshold is higher than 7.5, the tracking accuracy is relatively stable for CMC1-3 and WT sequences. This value is 10 for CMC4-5 sequences. Note that we observe the cost if a bounding box detection is correctly assigned to a track is between 4 and 8 for CMC1-3 and WT sequences. This cost is between 4 and 12 in CMC4-5 sequences due to the jumping and falling objects. We observe that the performance of the pose estimation follows the trend of the tracking performance in this study. Further, the trends on MultiviewX dataset are similar to those on WT.

The inferior performance at low gating thresholds is because when the threshold is lower than the cost of the correct assignment, the assignment algorithm incorrectly declares many tracks as miss-detected, and the unassigned detections are used to initialize false tracks. In contrast, if the threshold is high, false positive detections are assigned to miss-detected tracks or false tracks, which leads to a decrease in performance, as seen in CMC3 and CMC5.


\section{Conclusions}\label{conclusion}
We have proposed a fast online 3D multi-object tracking and pose estimation algorithm using 2D detections from monocular cameras. Our algorithm is based on a drastic approximation of the Bayes-optimal multi-sensor generalized labeled multi-Bernoulli filter. In particular, we determine the most suitable term in the multi-object filtering density through the track initialization/termination process and resolve multi-camera data association with a series of independent 2D linear assignments, admitting a low-cost
implementation. Experimental results demonstrate that our algorithm, which provides trajectories and poses in 3D world frame, is much faster than the state-of-the-art methods without compromising performance, even when cameras are reconfigured on-the-fly.

\section*{Acknowledgements}
The authors would like to acknowledge Professor Ba-Ngu Vo (ba-ngu.vo@curtin.edu.au ) for his invaluable discussions during the drafting of this manuscript. This work was supported in part by Culture, Sports and Tourism R\&D Program through the Korea Creative Content Agency grant funded by the Korea government (MCST) in 2024 (R2022060001), and GIST-MIT Research Collaboration Project.

\bibliographystyle{IEEEtran}
\bibliography{main}
\appendices
\setcounter{table}{0} 
\renewcommand{\thetable}{A\arabic{table}}
\setcounter{algocf}{0} 
\renewcommand{\thealgocf}{A\arabic{algocf}}

\twocolumn[{%
 \centering
 \Large Appendices for ``Fast Online 3D Multi-Camera Multi-Object Tracking and Pose Estimation" \\[0.5em]
 \normalsize Linh Van Ma, Tran Thien Dat Nguyen, and Moongu Jeon\\[1em]
}]


\section{Pseudocodes for Important Sub-Routines\label{sec:app-pseudocode}}
In this appendix, we provide a series of pseudocodes for the important sub-routines used in Algorithm \ref{alg:sf3dmot} in the main text.  The list of constants and notation is given in Table \ref{fig:pseudo-notations}.

\begin{table}[h!]
\fontsize{10}{8}
\caption{List of constants and notations used in the pseudocodes.\label{fig:pseudo-notations}}
\small
\begin{tabular}{c>{\raggedright}p{0.75\columnwidth}}
\toprule 
Notations & Meanings\tabularnewline
\midrule
\midrule 
$v$ & Number of cameras.\tabularnewline
$\mathbf{M}^{(c)}$ & Camera matrix of camera $c$.\tabularnewline
$\mathbf{R}^{(c)}_b$ & Covariance matrix of the bounding box detection noise of camera $c$.\tabularnewline
$\mathbf{R}^{(c)}_k$ & Covariance matrix of the keypoint detection noise of camera $c$.\tabularnewline
$\kappa$ & Secondary scaling factor of the unscented transform, set to 2.\tabularnewline
$\alpha$ & Sigma point spreading factor of the unscented transform, set to 1.\tabularnewline
$\beta$ & Prior knowledge factor of the unscented transform, set to 2.\tabularnewline
$\tau_{G}$ & Gating threshold for ground plane position.\tabularnewline
$\tau_{C}$ & Gating threshold for assignment cost.\tabularnewline
\bottomrule
\end{tabular}
\end{table}

\begin{algorithm}[h!]
\DontPrintSemicolon
\nonl\textbf{Input}: $b$, $\mu^{(s)}$, $\mathbf{P}^{(s)}$, $c$.  \;
\nonl\textbf{Output}: $q$, $\mu_u^{(s)}$, $\mathbf{P}_u^{(s)}$. \;
$L\leftarrow9$ \;
$n\leftarrow 2L+1$ \;
$(\mathcal{X},w_m,w_c)\leftarrow$\funcfont{UnscentedTransform}$(\mu^{(s)},\mathbf{P}^{(s)})$ (Algorithm \ref{alg:ut}) \;
\For{$i=1:n$}{
    $\mathcal{Y}[i]\leftarrow$\funcfont{3Dto2DBBox}$(\mathcal{X}[i],c)$ (Algorithm \ref{alg:genmsobservation})
}
$\mu_d\leftarrow b-\sum_{i=1}^{n}w_{m}[i]\mathcal{Y}[i]$ \;
$\mathbf{P}_d\leftarrow\sum_{i=1}^{n}w_{c}[i](\mathcal{Y}[i]-\mu_d)(\mathcal{Y}[i]-\mu_d)^T$ \;
$\mathbf{P}_s\leftarrow\sum_{i=1}^{n}w_{c}[i](\mathcal{X}[i]-\mu)(\mathcal{Y}[i]-\mu_d)^T$ \;
$\mathbf{K}\leftarrow\mathbf{P}_d\mathbf{P}_s^{-1}$ \;
$\mu_u^{(s)}\leftarrow\mu^{(s)}+\mathbf{K}\mu_d$ \;
$\mathbf{P}_u^{(s)}\leftarrow \mathbf{P}^{(s)}-\mathbf{P}_d\mathbf{P}_s^{-1}\mathbf{P}_s$ \;
$q\leftarrow$\funcfont{GaussianPDF}$(b,\mu_u^{(s)},\mathbf{R}^{(c)}_b+\mathbf{P}_u^{(s)})$ \;
\caption{\funcfont{UKFUpdateKS} \label{alg:ukfupdatepersensor}}
\end{algorithm}

\begin{algorithm}[h!]
\DontPrintSemicolon
\nonl\textbf{Input}: $b$, $\mu$, $c$. \;
\nonl\textbf{Output}: $s$. \;
$x\leftarrow$$\begin{bmatrix} 
                \mu[1]\\
                \mu[2] \\
                \end{bmatrix}$\;
$y\leftarrow$\funcfont{2DBBoxtoGroundPlane}$(b,c)$ (Algorithm \ref{alg:2DtoGP}) \;
\If{$||\thinspace x-y\thinspace ||_2>\tau_G$}{
$s\leftarrow$ \funcfont{false} \;
}
\Else{$s\leftarrow$ \funcfont{true}}
\caption{\funcfont{DetectionGating}. \label{alg:det-gating}}
\end{algorithm}

\begin{algorithm}[h!]
\DontPrintSemicolon
\nonl\textbf{Input}: $b$, $c$. \;
\nonl\textbf{Output}: $g$. \;
$f\leftarrow$$\begin{bmatrix} 
b[1]+\frac{\exp(b[3])}{2}\\
b[2]+\exp(b[4])  \\
\end{bmatrix}$ \;
$f\leftarrow\Big(\mathbf{M}^{(c)}[:,[1,2,3]]\Big)^{-1}f$ \;
$f[1]\leftarrow f[1]/f[3]$ \;
$f[2]\leftarrow f[2]/f[3]$\;
$g\leftarrow$$\begin{bmatrix} 
                f[1]\\
                f[2] \\
                \end{bmatrix}$\;
\caption{\funcfont{2DBBoxtoGroundPlane}. \label{alg:2DtoGP}}
\end{algorithm}

  


\begin{algorithm}[h!]
\DontPrintSemicolon
\nonl\textbf{Input}: $k$, $\mu^{(p)}$, $\mathbf{P}^{(p)}$, $c$, $\mathbf{M}^{(c)}$.  \;
\nonl\textbf{Output}: $\mu_u^{(p)}$, $\mathbf{P}_u^{(p)}$. \;
$[\mu^{(p)}_{i}]_{i=1:P}\leftarrow$\funcfont{ListKeyPointMean}$(\mu^{(p)})$ \;
$[\mathbf{P}^{(p)}_{i}]_{i=1:P}\leftarrow$\funcfont{ListKeyPointCovariance}$(\mathbf{P}^{(p)})$ \;
$L\leftarrow6$;
$n\leftarrow 2L+1$ \;
\For{$i=1:P$}{
        \If{$k_i\neq\overline{\infty}$}{$(\mathcal{X},w_{m},w_{c})\leftarrow$\funcfont{UnscentedTransform}$(\mu^{(p)}_{i},\mathbf{P}^{(p)}_{i})$ (Algorithm \ref{alg:ut}) \;
        \For{$j=1:n$}{
            $\mathcal{Y}[j]\leftarrow \mathbf{M}^{(c)} \mathcal{X}[j]$ \;
            $\mathcal{Y}[j]\leftarrow \left[\frac{\mathcal{Y}[j, 0]}{\mathcal{Y}[j, 2]}, \frac{\mathcal{Y}[j, 1]}{\mathcal{Y}[j, 2]}\right]$ \;
        }
        $\mu_d\leftarrow k_i - \sum_{j=1}^{n}w_{m}[j]\mathcal{Y}[j]$ \;
        $\mathbf{P}_d\leftarrow\sum_{j=1}^{n}w_{c}[j](\mathcal{Y}[j]-\mu_d)(\mathcal{Y}[j]-\mu_d)^T$ \;
        $\mathbf{P}_s\leftarrow\sum_{j=1}^{n}w_{c}[j](\mathcal{X}[j]-\mu_d)(\mathcal{Y}[j]-\mu_d)^T$ \;
        $\mathbf{K}\leftarrow\mathbf{P}_d\mathbf{P}_s^{-1}$ \;
        $\mu_{i}^{(p)}\leftarrow\mu_{i}^{(p)}+\mathbf{K}\mu_d$ \;
        $\mathbf{P}_{i}^{(p)}\leftarrow \mathbf{P}_{i}^{(p)}-\mathbf{P}_d\mathbf{P}_s^{-1}\mathbf{P}_s$ \;}
}
$\mu_{u}^{(p)}\leftarrow$\funcfont{KeypointMean}$([\mu^{(p)}_{i}]_{i=1:P})$ \;
$\mathbf{P}^{(p)}_{u}\leftarrow$\funcfont{KeypointCovariance}$([\mathbf{P}^{(p)}_{i}]_{i=1:P})$ \;
\caption{\funcfont{UKFUpdateKP} \label{alg:ukfupdatepersensor_kp}}
\end{algorithm}

\begin{algorithm}[h!]
\DontPrintSemicolon
\nonl\textbf{Input}: $\mu$, $\mathbf{P}$. \;
\nonl\textbf{Output}: $\mathcal{X}$, $w_m$,  $w_c$. \;
$L\leftarrow $\funcfont{Dimension}($\mu$ ) \;
$n\leftarrow 2L+1$ \;
$\lambda\leftarrow\alpha^2(L+\kappa)-L$ \
$\mathcal{X}[1]\leftarrow\mu$ \;
$w_{m}[1]\leftarrow\frac{\lambda}{L+\lambda}$ \;
$w_{c}[1]\leftarrow\frac{\lambda}{L+\lambda}+1-\alpha^2+\beta$ \;
$\mathbf{P}_{sq}\leftarrow\sqrt{(L+\lambda)\mathbf{P}}$ \;
\For{$i=2:L+1$}{
    $\mathcal{X}[i]\leftarrow\mu+\mathbf{P}_{sq}[i,:]$
}
\For{$i=L+2:2L+1$}{
    $\mathcal{X}[i]\leftarrow\mu-\mathbf{P}_{sq}[i-L,:]$
}
\For{$i=2:2L+1$}{
    $w_{m}[i]\leftarrow\frac{1}{2(L+\lambda)}$ \;
    $w_{c}[i]\leftarrow\frac{1}{2(L+\lambda)}$
}
\caption{\funcfont{UnscentedTransform} \label{alg:ut}}
\end{algorithm}

\begin{algorithm}[h!]
\DontPrintSemicolon
\nonl\textbf{Input}: $x$, $c$, $\mathbf{M}^{(c)}$ (camera matrix of camera $c$). \;
\nonl\textbf{Output}: $b$ . \;

$\rho\leftarrow x[1:3]$ \quad\funcfont{/* $x[n:m]$: the $n^{th}$ to $m^{th}$ components of $x$ */} \;
$s\leftarrow$ exp$(x[7:9])$ \;
$\mathbf{A}\leftarrow$$\begin{bmatrix} 
    \rho[1] + s[1] & \rho[2] & \rho[3] & 1 \\
    \rho[1] - s[1] & \rho[2] & \rho[3] & 1 \\
    \rho[1]      & \rho[2]+s[2] & \rho[3] & 1 \\
    \rho[1]      & \rho[2]-s[2] & \rho[3] & 1 \\
    \rho[1]      & \rho[2] & \rho[3]+s[3] & 1 \\
    \rho[1]      &  \rho[2] & \rho[3]-s[3] & 1 \\
    \end{bmatrix}$ \;
$\mathbf{P}\leftarrow\mathbf{M}^{(c)}\mathbf{A}^{T}$ \;
$\zeta\leftarrow \mathbf{0}_{2\times 6}$ \;
$\zeta[1,:]\leftarrow\mathbf{P}[1,:]\oslash\mathbf{P}[3,:]$ \funcfont{/* $\zeta[n,:]$: all elements in row $n^{th}$ of $\zeta$ */} \;
$\zeta[2,:]\leftarrow\mathbf{P}[2,:]\oslash\mathbf{P}[3,:]$ \funcfont{/* $\oslash$: element-wise division*/} \;
$l\leftarrow$\funcfont{min}$(\zeta[1, :])$    \quad\funcfont{/* Left */} \;
$t\leftarrow$\funcfont{min}$(\zeta[2, :])$    \quad\funcfont{/* Top */} \;
$r\leftarrow$\funcfont{max}$(\zeta[1, :])$    \quad\funcfont{/* Right */} \;
$o\leftarrow$\funcfont{max}$(\zeta[2, :])$    \quad\funcfont{/* Bottom */} \;
$b\leftarrow[l,t,\text{log}(r-l),\text{log}(o-t)]^{T}$ \;
\caption{\funcfont{3Dto2DBBox}.\label{alg:genmsobservation}}
\end{algorithm}

\section{Tracking Performance Comparison with 3D-Training-Based Methods\label{subsec:benchmarkideal}}

In this appendix, using WILDTRACK (WT) dataset, we evaluate our algorithm against 3D-training-based methods. Note that the 3D-training-based methods use a portion of the ground truth as training data. In particular, we use classical multi-object tracking filters including the single-sensor generalized labeled multi-Bernoulli (GLMB) \cite{vo2013labeled}, multi-hypothesis tracking (MHT), joint probabilistic data association (JPDA), global nearest neighbor (GNN) filters \cite{blackman1999design} to process 3D detections obtained from 3DROM detector \cite{Qiu20223DRO}. We also test the KSP-ptracker (K.p.) \cite{chavdarova2018wildtrack} using 3D detections from
DeepOcclusion (DeepOcc.) detector \cite{baque2017deep}, LGMP \cite{nguyen2022lmgp} and EarlyBird \cite{teepe2024earlybird} 3D trackers. We note that 3DROM, DeepOcc. and LGMP models are trained on 90\% of the WT dataset and evaluated on the remaining 10\%. The EarlyBird tracker was trained on 10-90\% of the WT dataset. The YOLOX 2D detector used in our method is not trained on any portion of the WT dataset or 3D data.

\begin{table}[h!]
\centering{}
\global\long\def\arraystretch{1.3}%
 \caption{Detection quality on the WT dataset (`$\ast$' indicates 3D) in terms of multi-object detection accuracy (MODA), multi-object detection precision (MODP) scores with their recall (Rcll) and precision (Prcn) components.\label{tbl:detquality-WT}}
\scriptsize
\setlength{\tabcolsep}{2mm}{
\begin{tabular}{|l|c|c|c|c|}
\hline 
Detector & MODA$\uparrow$ & MODP$\uparrow$ & Rcll$\uparrow$ & Prcn$\uparrow$\tabularnewline
\hline 
\hline 
3DROM{*} & 93.5 & 75.9 & 96.2 & 97.2\tabularnewline
\hline 
YOLOX & 38.09 & 67.02 & 76.22 & 66.66\tabularnewline
\hline 
\end{tabular}}
\end{table}

Table \ref{tbl:detquality-WT}, which presents the detection quality on WT dataset, shows that the 3DROM detector achieves almost perfect results. Note that the reported YOLOX detection quality is averaged over the results from all cameras. We observe that YOLOX detections are much worse than 3DROM detections. Tracking results are reported in Table \ref{tbl:wildtrack_3ddet-1} which shows that LMGP is the best method among the tested ones. In terms of IDF1, our method is better than classical single-sensor MOT filters with 3DROM detections except for the GLMB filter. In general, our method is inferior to the recent LMGP and EarlyBird trackers, which are trained on a portion of the ground truth. Nonetheless, in terms of IDF1, our method is comparable with the EarlyBird tracker trained on 30\% of the ground truth and better than one that is trained on 10\% of the ground truth. The inferior performance of our method is due to the low-quality 2D detection results compared to the 3D-training-based detection. However, our method only requires generic 2D detection models, not scenario-dependent 3D detection/tracking models.




\begin{table}[h!]
\centering{}
\global\long\def\arraystretch{1.3}%
 \caption{Tracking performance of our algorithm
with YOLOX detector and 3D-training-based methods. The third column indicates the percentage of the WT dataset that is used for 3D training purposes (evaluation is performed over the remaining portion). The best result for each column is \underline{\textbf{bolded and underlined}}.\label{tbl:wildtrack_3ddet-1}}
\centering{}
\scriptsize
\setlength{\tabcolsep}{2mm}{
\begin{tabular}{|c|c|c|ccc|}
\hline 
Detector & Tracker & \%3D$\downarrow$ & MOTA$\uparrow$ & IDF1$\uparrow$ & OSPA\!$^{\texttt{(2)}}$\!$\downarrow$\tabularnewline
\hline 
\hline 
YOLOX & Ours & \underline{\textbf{0}} & 47.6  & 75.0  & 0.76\tabularnewline
\hline 
\multirow{4}{*}{3DROM} & GLMB & 90 & 81.6 & 86.4 & \underline{\textbf{0.19}}\tabularnewline
 & MHT & 90 & 51 & 50.2 & 0.31\tabularnewline
 & JPDA & 90 & 63 & 60.1 & 0.39\tabularnewline
 & GNN & 90 & 57.6 & 55.4 & 0.52\tabularnewline
\hline 
Deep Occ. & K.p. & 90 & 72.2 & 78.4 & 0.75\tabularnewline
\hline 
\multirow{4}{*}{EarlyBird} & \multirow{4}{*}{EarlyBird} & 90 & 90.4  &  93.3  &  0.27\tabularnewline
& & 50 &  84.0  &  85.4  &  0.48\tabularnewline
& & 30 &  74.1  &  76.4  &  0.60\tabularnewline
& & 10 &  62.3  &  67.8  &  0.71\tabularnewline
\hline
LMGP & LMGP & 90 & \underline{\textbf{97.1}}  &  \underline{\textbf{98.2}}  &  -\tabularnewline
\hline 
\end{tabular}}
\end{table}



\end{document}